\documentclass[10pt,twocolumn,letterpaper]{article}

\usepackage{cvpr}
\usepackage{times}
\usepackage{epsfig}
\usepackage{graphicx}
\usepackage{amsmath}
\usepackage{amssymb}
\usepackage{color}
\usepackage[T1]{fontenc}
\usepackage[latin9]{inputenc}
\usepackage{xcolor}
\usepackage{array}
\usepackage{verbatim}
\usepackage{units}
\usepackage{textcomp}
\usepackage{bbding}
\usepackage{url}
\usepackage{multirow}
\usepackage{times}
\usepackage{epsfig}
\usepackage{placeins}
\usepackage{colortbl}
\usepackage{rotating}
\definecolor{lightgray}{gray}{0.6}
\definecolor{verylightgray}{gray}{0.9}

\usepackage{caption} 
\usepackage{hyphenat}
\usepackage[british]{babel}
\usepackage{tablefootnote}
\usepackage{wrapfig}

\usepackage{enumitem}

\usepackage[breaklinks=true,bookmarks=false]{hyperref}

\cvprfinalcopy % *** Uncomment this line for the final submission

\def\cvprPaperID{0007} % *** Enter the CVPR Paper ID here

\makeatletter
\renewcommand{\paragraph}{%    
\@startsection{paragraph}{4}%  
{\z@}{0.5ex \@plus 1ex \@minus .2ex}{-0.5em}% 
{\normalfont \normalsize \bfseries}%  
}

 \let\originalparagraph\paragraph 
 \renewcommand{\paragraph}[2][.]{\originalparagraph{#2#1}}
 \makeatother

\begin{document}

%%%%%%%%% TITLE %%%%%%%%%
\vspace{-1em}
\title{Learning to Refine Human Pose Estimation}

\author{Mihai Fieraru \qquad Anna Khoreva \qquad Leonid Pishchulin \qquad Bernt Schiele \\
{\tt\small \{mfieraru, khoreva, leonid, schiele\}@mpi-inf.mpg.de}\\
Max Planck Institute for Informatics \\
Saarland Informatics Campus\\
Saarbr\"ucken, Germany
}

\maketitle
% \thispagestyle{empty}

%%%%%%%%% ABSTRACT %%%%%%%%%
\begin{abstract}
Multi-person pose estimation in images and videos is an important yet challenging task with many applications. Despite the large improvements in human pose estimation enabled by the development of convolutional neural networks, there still exist a lot of difficult cases where even the state-of-the-art models fail to correctly localize all body joints. This motivates the need for an additional refinement step that addresses these challenging cases and can be easily applied on top of any existing method. In this work, we introduce a pose refinement network (PoseRefiner) which takes as input both the image and a given pose estimate and learns to directly predict a refined pose by jointly reasoning about the input-output space. In order for the network to learn to refine incorrect body joint predictions, we employ a novel data augmentation scheme for training, where we model ``hard`` human pose cases. We evaluate our approach on four popular large-scale pose estimation benchmarks such as MPII Single- and Multi-Person Pose Estimation, PoseTrack Pose Estimation, and PoseTrack Pose Tracking, and report systematic improvement over the state of the art.

\end{abstract}

%%%%%%%%% BODY TEXT %%%%%%%%%

\section{\label{sec:Intro} Introduction}

The task of human pose estimation is to correctly localize and estimate body poses of all people in the scene. 
Human poses provide strong cues and have shown to be an effective representation for a variety of tasks such as activity recognition, motion capture, content retrieval
and social signal processing. Recently, human pose estimation performance has improved significantly due to the use of deep convolutional neural networks \cite{PapandreouZKTTB17,insafutdinov2016deepercut,yang2017learning,newell2017associative,Gler2018DensePoseDH} 
and availability of large-scale datasets \cite{lin2014microsoft,andriluka2017posetrack,andriluka20142d,LIPdataset}.
\begin{figure}
\vspace{-1em}
\begin{centering}

\hfill{}%
\begin{tabular}{c@{\hskip 0.01in}c@{\hskip 0.01in}c}

\includegraphics[width=0.31\columnwidth]{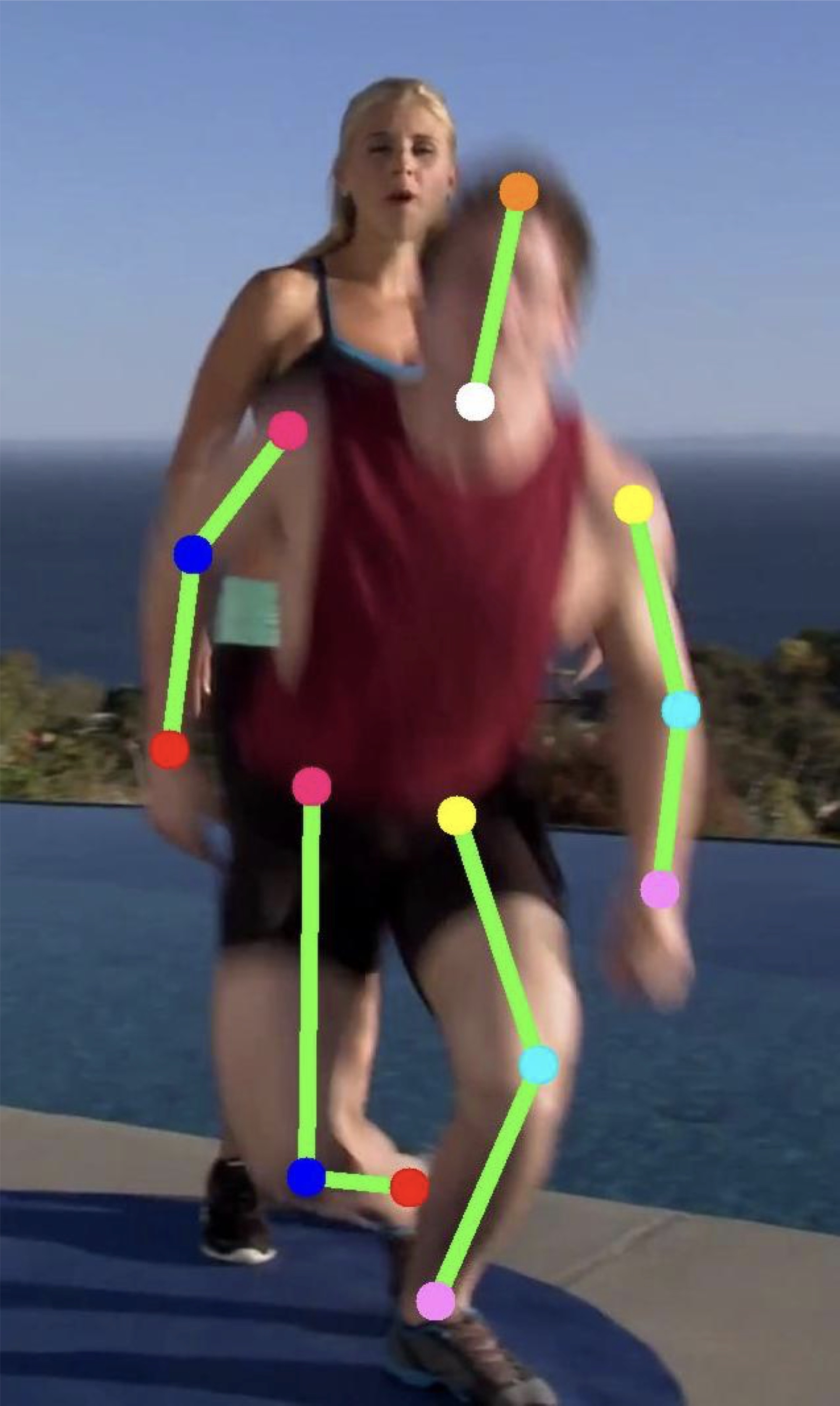} & {\footnotesize{}}
\includegraphics[width=0.31\columnwidth]{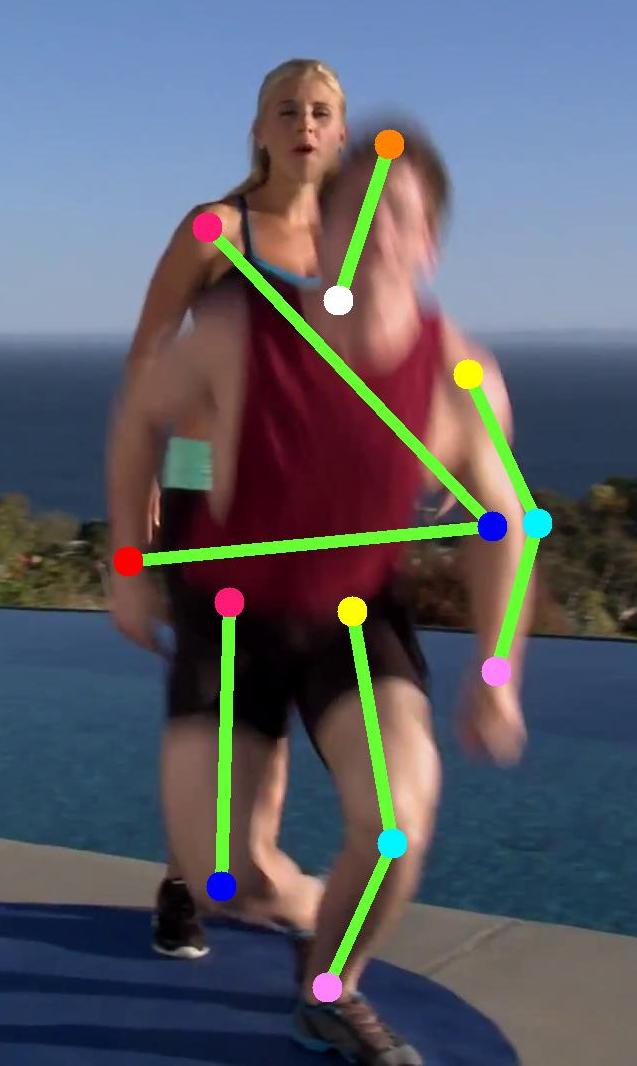} & {\footnotesize{}}
\includegraphics[width=0.31\columnwidth]{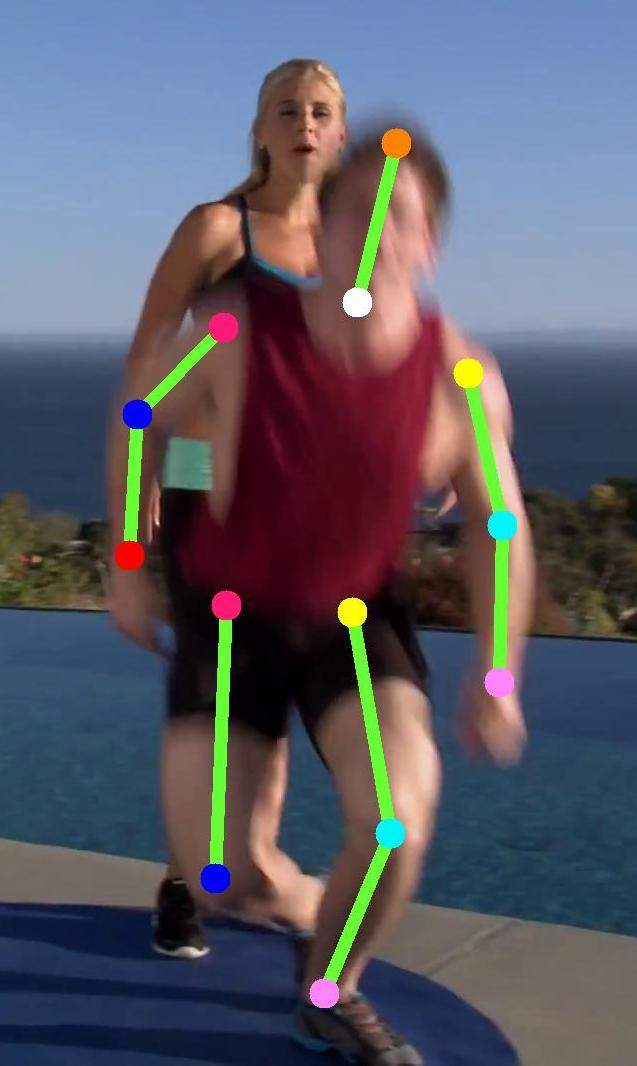} \tabularnewline

Ground Truth & {\footnotesize{}}
Initial & {\footnotesize{}}
Refined \tabularnewline
Pose & {\footnotesize{}}
Pose Estimate & {\footnotesize{}}
Output Pose \tabularnewline

\end{tabular}\hfill{}

\par\end{centering}
\caption{\label{fig:Teaser} 
Example of the proposed pose refinement. Starting from an image and an estimated body pose \textbf{(central)}, our refinement method \texttt{PoseRefiner} outputs a denoised body pose \textbf{(right)}. The system learns to fuse the appearance of the person and an estimation of its pose structure in order to better localize each body joint. It is trained to specifically target common errors of human pose estimation methods, e.g. merges of body joints of different people in close proximity and confusion between right/left joints.}
\vspace{-1em}
\end{figure}

Although great progress has been made, the problem remains far from being solved.
There still exist a lot of challenging cases, such as person-person occlusions, close proximity of similar looking people, rare body configurations, partial visibility of people and cluttered
backgrounds. Despite the great representational power, current deep learning-based approaches are not explicitly trained to address such hard cases and often output incorrect body pose predictions such as spurious body configurations, 
merges of body joints of different people, confusion between similarly looking left and right limbs, or missing joints.

In this work, we propose a novel human pose refinement approach that is explicitly trained to address such hard cases. Our simple yet effective pose refinement method can be applied on top of any body pose prediction computed by an arbitrary human pose estimation approach, and thus is complementary to current approaches \cite{yang2017learning,newell2017associative,cao2017realtime,insafutdinov2017arttrack}. As we demonstrate empirically, the proposed pose refinement allows to push the state of the art on several standard benchmarks of single- and multi-person pose estimation \cite{andriluka20142d,andriluka2017posetrack}, as well as articulated pose tracking \cite{andriluka2017posetrack}.
In more detail, given an RGB image and a body pose estimate, we aim to output a refined human pose by exploiting the dependencies between the image and the inherent structure of the provided body pose (see Figure~\ref{fig:Teaser}). This makes it easier 
for the network to identify what is wrong with input prediction and find a way to refine it. We employ a fully convolutional ResNet-based architecture and propose an elaborate data augmentation scheme for training. 
To model challenging cases, we propose a novel data augmentation procedure that allows to synthesize possible input poses and make the network learn to identify the erroneous body joint predictions and to refine them. We refer to the proposed approach as {\texttt{PoseRefiner}}.

We evaluate our approach on four human pose estimation benchmarks, namely MPII Single Person \cite{andriluka20142d}, MPII Multi-Person \cite{andriluka20142d}, PoseTrack Multi-Person Pose Estimation \cite{andriluka2017posetrack}, and PoseTrack Multi-Person Pose Tracking \cite{andriluka2017posetrack}.
We report consistent improvement after applying the proposed refinement network to pose predictions given by various state-of-the-art approaches 
\cite{ml_lab,girdhar2017detect,insafutdinov2017arttrack,jin2017towards,gkioxari2016chained,yang2017learning,newell2017associative,varadarajan2017greedy,cao2017realtime,carreira2016human}
across different datasets and tasks, showing the effectiveness and generality of the proposed framework.
With our refinement network, we improve the best reported results for multi-person pose estimation and pose tracking on MPII Human Pose and PoseTrack datasets.

In summary, our contributions are as follows:
\begin{itemize}
 \item We introduce an effective post-processing technique for body joint refinement in human pose estimation tasks, that works on top of any
existing human body pose estimation approach. Our proposed pose refinement network is efficient due to its feed-forward architecture, simple and end-to-end trainable.
 \item We propose a training data augmentation scheme for error correction, which enables the network to identify the erroneous body joint predictions and to learn a way to refine them.
 \item We show that our refinement model allows to systematically improve over various state-of-the-art methods and achieve top performing results on four different benchmarks.
 \end{itemize}

The rest of the paper is organized as follows. Section \ref{sec:Related-work} provides an overview of the related work and positions the proposed approach with respect to earlier work. 
Section \ref{sec:Method} describes the proposed pose refinement network and data augmentation for error correction of human body pose estimation. 
Experimental results are presented in Section \ref{sec:Results}. Section \ref{sec:Conclusion} concludes the paper.

\section{\label{sec:Related-work} Related Work}

Our proposed approach is related to previous work on single- and multi-person pose estimation, articulated tracking as well as refinement/error correction methods, as described
next.

\paragraph{Single-person pose estimation}

Classical methods \cite{Gkioxari2013ArticulatedPE, Yang2011ArticulatedPE, Andriluka2009PictorialSR, Sapp2010AdaptivePP, Dantone2013HumanPE, Pishchulin2013PoseletCP} formulate single person pose estimation 
as a pictorial structure or graphical model problem and predict body joint locations using only hand-designed features.
More recent methods \cite{Toshev2014DeepPoseHP,Tompson2014JointTO, Wei2016ConvolutionalPM, Newell2016StackedHN, Lifshitz2016HumanPE, yang2017learning} rely on localizing body joints by employing convolutional neural networks (CNNs),
which contributed to large improvement in human pose estimation.
\cite{Toshev2014DeepPoseHP} directly predicts joint coordinates via a cascade of CNN pose regressors, while further improvement in the performance is achieved by 
predicting heatmaps of each body joint \cite{Tompson2014JointTO, Lifshitz2016HumanPE} and using very deep CNNs with multi-stage architectures \cite{Wei2016ConvolutionalPM}. 
Our method is complementary to current approaches, as it is able to use their predictions as input and further improve their results, see Section \ref{sec:Results-single-pose} for details.

\paragraph{Multi-person pose estimation}

Compared to single person pose estimation, multi-person pose estimation requires
parsing of the full body poses of all people in the scene and is a much more challenging task due to occlusions, various articulations and interactions between people.
Multi-person pose estimation methods can be grouped into two types: top-down and bottom-up approaches.

Top-down approaches \cite{PapandreouZKTTB17,Huang2017ACN, he2017maskrcnn,fang2017rmpe,Iqbal_CVPR2017, kposelets} employ a person detector and then perform single-person pose
estimation for each detected person. These methods highly depend on the reliability of the person detector and are known to have trouble recovering poses of people in close proximity to each other and/or with overlapping body parts. 
Thus, the output predictions of top-down methods might benefit from an additional refinement step proposed in this work.

Bottom-up methods \cite{cao2017realtime,pishchulin2016deepcut,insafutdinov2016deepercut,newell2017associative, Ladicky2013HumanPE} first predict
all body joints and then group them into full poses of different people. Instead of applying person detection, these methods rely on 
context information and inter body joint relationships. However, modeling the joint relationships might not be as reliable, causing mistakes like failure to disambiguate poses of different people 
or grouping body parts of the same person into different clusters. 
Our refinement approach can particularly help in this scenario, as it re-estimates joint locations by taking the structure of the body pose into account.
\begin{figure*}
\vspace{-3em}
\begin{centering}
\includegraphics[width=1.0\textwidth]{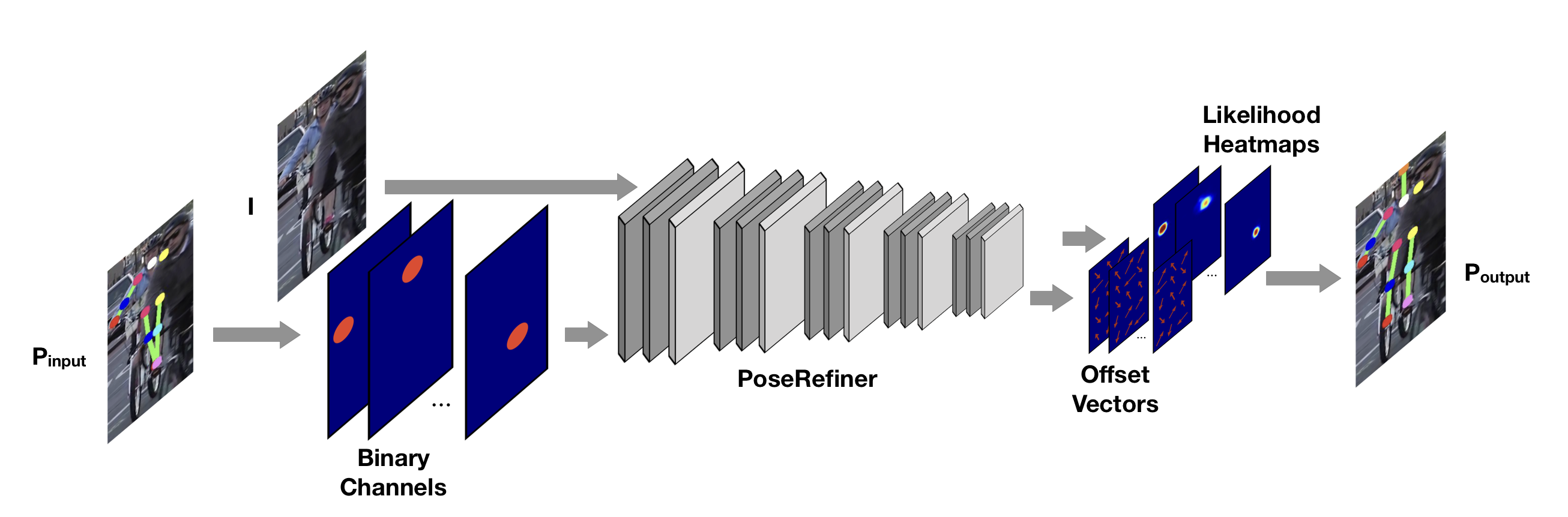}
\par\end{centering}
\caption{\label{fig:System} The overview of our \texttt{PoseRefiner} system. We take as input an image $I$ and an initial estimate of a person body pose $P_{input}$. The input pose is encoded as $n$ binary channels, where $n$ is the number of joints, which are stacked together with the image RGB channels and used as an input to a fully convolutional network. The network learns to predict likelihood heatmaps for each joint type, as well as offset vectors to recover from the downscaled spatial resolution. The output pose $P_{output}$ is a refined estimate of the initial input.}
\vspace{-0.5em}
\end{figure*}

\paragraph{Articulated pose tracking}

Most articulated pose tracking methods rely on a two-stage framework, which first employs a per-frame pose estimator and then smooths the predictions over time.
\cite{SongWVH17} proposes a model combining a CNN and a CRF to jointly optimize per-frame predictions with the CRF, smoothing the predictions over space and time.
\cite{Iqbal_CVPR2017, insafutdinov2017arttrack,jin2017towards} employ a bottom-up strategy, they first detect body joints of all people in all frames
of the video and then an integer program optimization is solved grouping joints into people over time.
\cite{girdhar2017detect} applies 3D Mask R-CNN \cite{he2017maskrcnn} over short video clips, producing a tubelet with body joints per person, and
then performs a lightweight optimization to link the predictions over time.
\cite{jin2017towards} extends the work of \cite{cao2017realtime} by rethinking the network architecture and developing a redundant part affinity fields (PAFs) mechanism, while \cite{ICG} 
employs a geometric tracker to match the predicted poses frame-by-frame. All of these approaches heavily rely on accurate pose estimation in a single frame. We show in Section~\ref{sec:Results-pose-tracking} that by refining the initial pose hypothesis in individual frames we are able to significantly improve pose tracking over time.

\paragraph{Refinement/error correction}

Another group of work \cite{carreira2016human,gkioxari2016chained,Belagiannis2017RecurrentHP,Neverova2017MassDN,Li2016IterativeIS} aims to refine labels from the initial estimate by jointly reasoning about input-output space.
\cite{carreira2016human} proposes to iteratively estimate residual corrections which are added to the initial prediction. In a similar spirit \cite{gkioxari2016chained,Belagiannis2017RecurrentHP} use RNN-like architectures
to sequentially refine the results and \cite{Pang2017CascadeRL,Chen2017CascadedPN,Li2016IterativeIS} employ cascade CNNs with refinement stages.
\cite{Gidaris2017DetectRR} decomposes the label improvement into three stages: first detecting the errors in the initial labels, replacing the incorrect
labels with new ones and refining the labels by predicting residual corrections.
Likewise \cite{Huang2017ErrorCF} employs a parallel architecture that propagates correct labels to nearby pixels and replaces the erroneous predictions with refined ones, then fuses the intermediate
results to obtain a final prediction.
In contrast to these methods, our proposed refinement network is much simpler as it learns to directly predict the refined labels from the initial estimate using a simple feed-forward fully convolutional network.

\section{\label{sec:Method}Method}
\begin{figure*}
\vspace{-1em}
\begin{centering}

\hfill{}%
\begin{tabular}{c@{\hskip 0.15in}c@{\hskip 0.15in}c@{\hskip 0.15in}c@{\hskip 0.15in}c}

\includegraphics[width=0.17\textwidth]{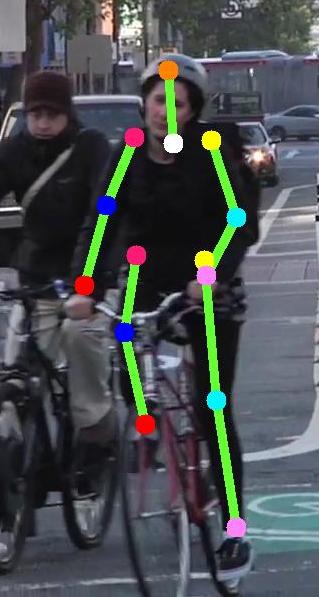} & {\footnotesize{}}
\includegraphics[width=0.17\textwidth]{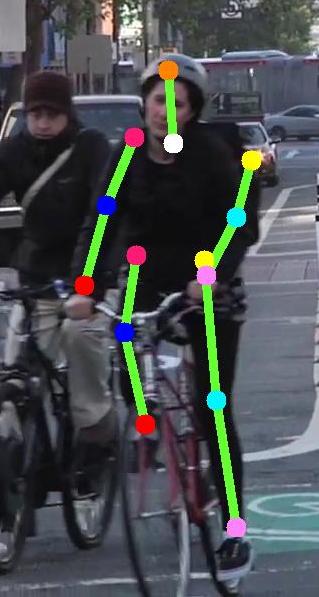} & {\footnotesize{}}
\includegraphics[width=0.17\textwidth]{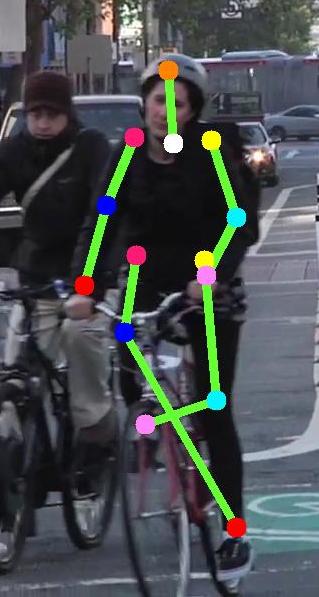} & {\footnotesize{}}
\includegraphics[width=0.17\textwidth]{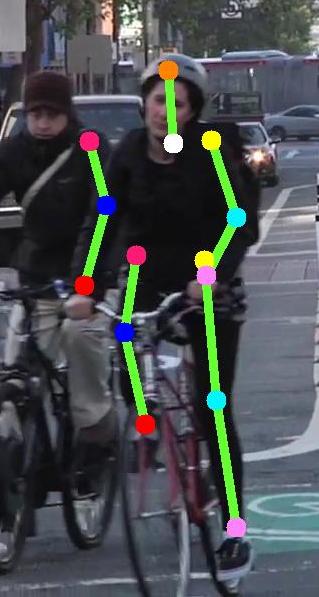} & {\footnotesize{}}
\includegraphics[width=0.17\textwidth]{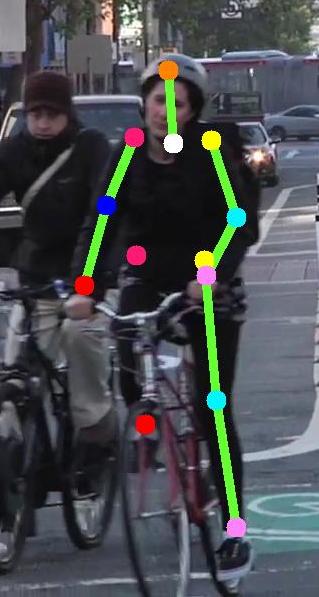} \tabularnewline

$P_{output}$ & {\footnotesize{}}
$P_{input}$ = \textbf{T1}$(P_{output})$ & {\footnotesize{}}
$P_{input}$ = \textbf{T2}$(P_{output})$ & {\footnotesize{}}
$P_{input}$ = \textbf{T3}$(P_{output})$ & {\footnotesize{}}
$P_{input}$ = \textbf{T4}$(P_{output})$ \tabularnewline

\end{tabular}\hfill{}

\par\end{centering}
\caption{\label{fig:synth} Examples of the proposed data synthesis for training. Starting from the ground truth $P_{output}$, we synthesize the initial pose estimate $P_{input}$ to mimic the most common errors of pose estimators. For visualization purposes, we only illustrate one transformation at a time: \textbf{T1} shifts the left shoulder (yellow), \textbf{T2} switches the left ankle (pink) with the right ankle (red), \textbf{T3} replaces the right shoulder (fuchsia) by the left shoulder of the neighboring person and \textbf{T4} removes the right knee.}
\vspace{0em}
\end{figure*}

In this section, we describe our approach - {\texttt{PoseRefiner}} - in detail.
We propose a pose refinement network which takes as input both an RGB image and a body pose estimate and aims to refine the initial prediction by jointly reasoning about the input and output space (see Figure \ref{fig:System}).
Exploiting the dependencies between the image and the predicted body pose makes it easier for the model to identify the errors in the initial estimate and how to refine them.
For the network to be able to learn to correct the erroneous body joint predictions we employ a training data augmentation scheme, modeled to generate the most common failure cases of human pose estimators.
This yields a model that is able to refine a human pose estimate derived from different pose estimation approaches and allows to achieve state-of-the-art results on the
challenging MPII Human Pose and PoseTrack benchmarks.

\paragraph{Approach}
We approach the refinement of pose estimation as a system on its own, which can be easily used as a post-processing step following any keypoint prediction task. 
Although there can be multiple estimated people poses in an image, we apply the refinement process on a per-person level. Given an estimated person pose, we initially rescale and crop around it to obtain a reference input. 
We then forward this obtained image $I$ and pose estimate $P_{input}$ as an input to a fully convolutional neural network $f$, modeled to compute a refined pose prediction $P_{output}$.

Formally, we refine an initial pose estimate $P_{input}$ as: $P_{output} = f(I, P_{input})$, where:

\begin{itemize}
\item $f$ is the \textbf{\texttt{PoseRefiner}}, the function to be learned. Since the output of this function is a single person pose, we model it using a fully convolutional network designed for single-person pose estimation.
\item $I$ is the original image in RGB format.
\item $P_{input}$ is the initial body pose, which needs to be refined. It is concatenated with the RGB image as $n$ additional channels, where $n$ is the number of body joints.
\item $P_{output}$ is the refined pose, in the form of $n$ channels. 
\end{itemize}

Both $P_{input}$ and $P_{output}$ are encoded using one binary channel for each body joint. \\

%%%%%%%%%%%%%% Architecture %%%%%%%%%%%%%%%%%
\paragraph{Architecture} 
We adopt the design choices that were shown successful in architectures with strong body joint detectors. 
As network architecture, we employ the ResNet-101 \cite{he2016deep} backbone converted to a fully convolutional mode with stride of 8 px. 
Although the ResNet-101 network is designed to accept as input only $3$ (RGB) channels, it can be easily extended to accept additional body joint channels by increasing the depth of the filters of the first convolutional layer (from $3$ to $3 + n$), 
where $n$ is the number of body joints.

Following \cite{pishchulin2016deepcut,insafutdinov2016deepercut}, we train the network to predict two types of output: likelihood heatmaps of each body joint and offsets from the locations on the heatmap grid to the ground truth joint locations. 
Likelihood heatmaps for each joint type are trained using sigmoid activations and cross entropy loss function. The shape of the output heatmaps is 8 times smaller in each spatial dimension than the shape of the input, due to the 8 pixel stride of the network. 
Hence to recover from the lost resolution, we learn to predict offset vectors from every heatmap location to the ground truth joint coordinate by regressing displacements $(\Delta x, \Delta y)$ using mean squared error. 

At test time, every pixel location in each likelihood heatmap indicates the probability of presence of the particular joint at that coordinate. 
The pixel with the highest confidence in each likelihood heatmap is selected as the rough downscaled joint coordinate. 
The final coordinate is obtained by adding the offset vector $(\Delta x, \Delta y)$ to the upscaled joint location predicted at the lower resolution. \\

%%%%%%%%%%%%%% Training Data Synthesis %%%%%%%%%%%%%%%%%
\paragraph{Training Data Synthesis} 

To train the network $f$, we need to have access to ground truth triplets $(I, P_{input}, P_{output})$. 
While $(I, P_{output})$ pairs are already available in large scale pose estimation datasets, we propose to synthesize $P_{input}$ to mimic the most common failure cases of human pose estimators. 
The goal is for the model to be able to refine initial estimates and  become robust to the "hard" body pose cases.
In essence, $P_{input}$ is a noisy version of $P_{output}$, which we synthesize from the ground truth by applying the following transformations (visualized in Figure~\ref{fig:synth}):
\begin{figure*}
\begin{centering}
\setlength{\tabcolsep}{0.1em}
\renewcommand{\arraystretch}{0}
\par\end{centering}
\begin{centering}
\vspace{-1em}
\hfill{}%
\begin{tabular}{c@{\hskip 0.05in}c@{\hskip 0.05in}c@{\hskip 0.05in}c@{\hskip 0.05in}c@{\hskip 0.05in}c@{\hskip 0.05in}c@{\hskip 0.05in}c}

%% ROW 1 %%
\includegraphics[width=0.15\textwidth,height=0.19\textheight]{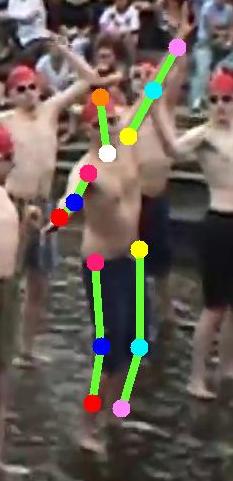} & {\footnotesize{}}
\includegraphics[width=0.15\textwidth,height=0.19\textheight]{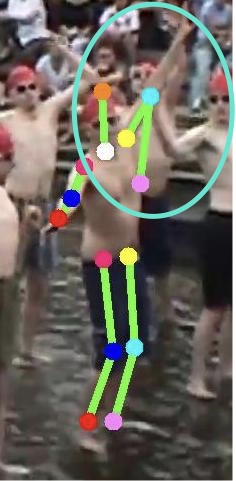} & {\footnotesize{}}
\includegraphics[width=0.15\textwidth,height=0.19\textheight]{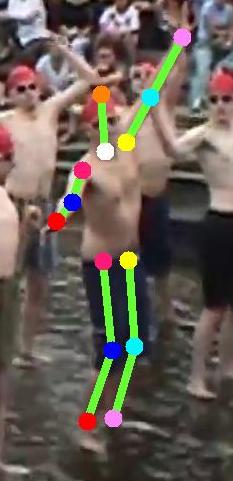} & {\footnotesize{}}
\includegraphics[width=0.15\textwidth,height=0.19\textheight]{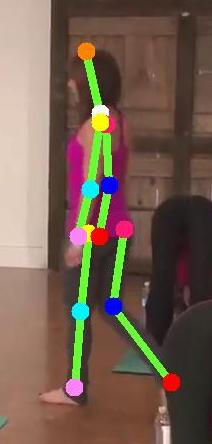} & {\footnotesize{}}
\includegraphics[width=0.15\textwidth,height=0.19\textheight]{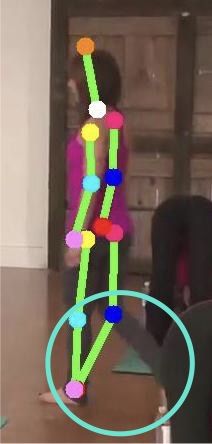} & {\footnotesize{}}
\includegraphics[width=0.15\textwidth,height=0.19\textheight]{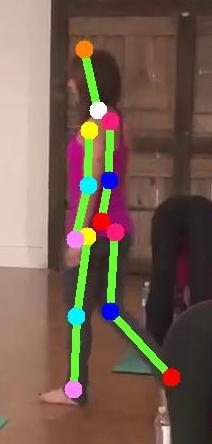} & {\footnotesize{}}\tabularnewline

GT Pose  & {\footnotesize{}} Initial Pose \cite{yang2017learning} & {\footnotesize{}} Refined Pose & GT Pose  & {\footnotesize{}} Initial Pose \cite{gkioxari2016chained}& {\footnotesize{}} Refined Pose \tabularnewline

%% ROW 2 %%
\includegraphics[width=0.15\textwidth,height=0.19\textheight]{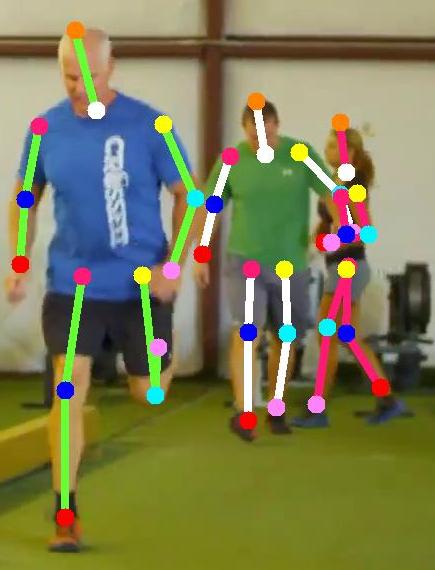} & {\footnotesize{}}
\includegraphics[width=0.15\textwidth,height=0.19\textheight]{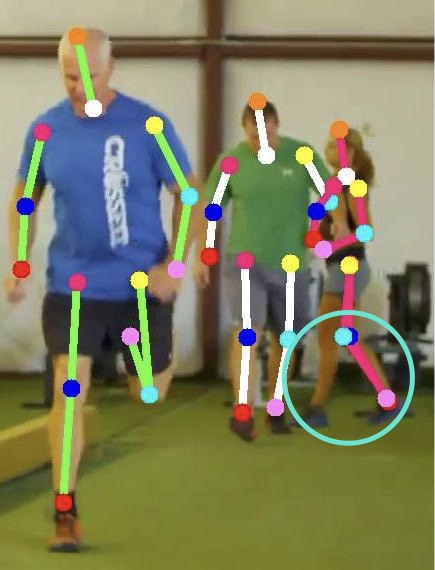} & {\footnotesize{}}
\includegraphics[width=0.15\textwidth,height=0.19\textheight]{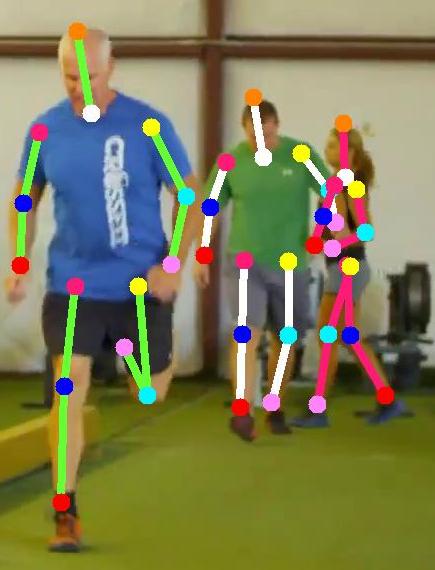} & {\footnotesize{}}
\includegraphics[width=0.15\textwidth,height=0.19\textheight]{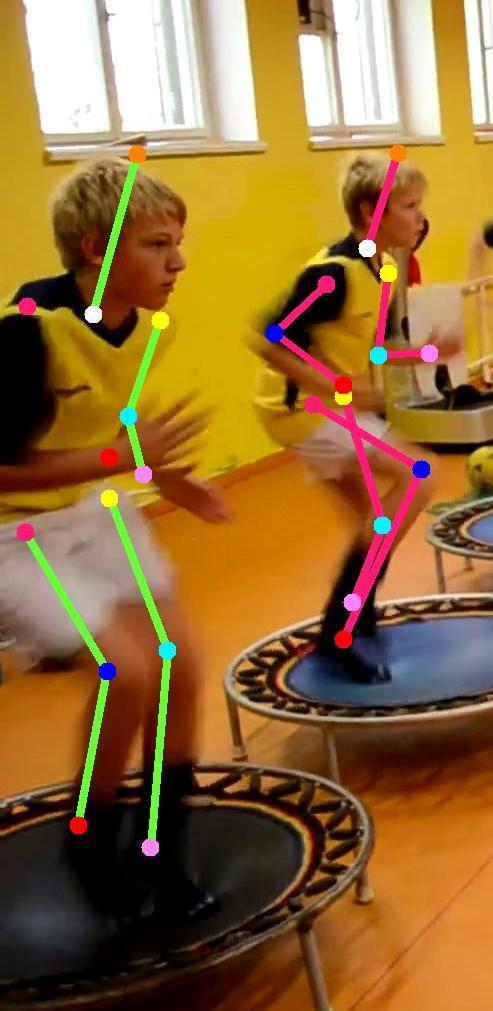} & {\footnotesize{}}
\includegraphics[width=0.15\textwidth,height=0.19\textheight]{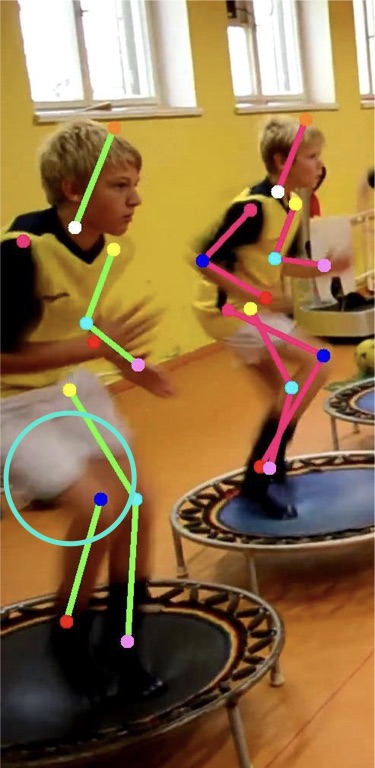} & {\footnotesize{}}
\includegraphics[width=0.15\textwidth,height=0.19\textheight]{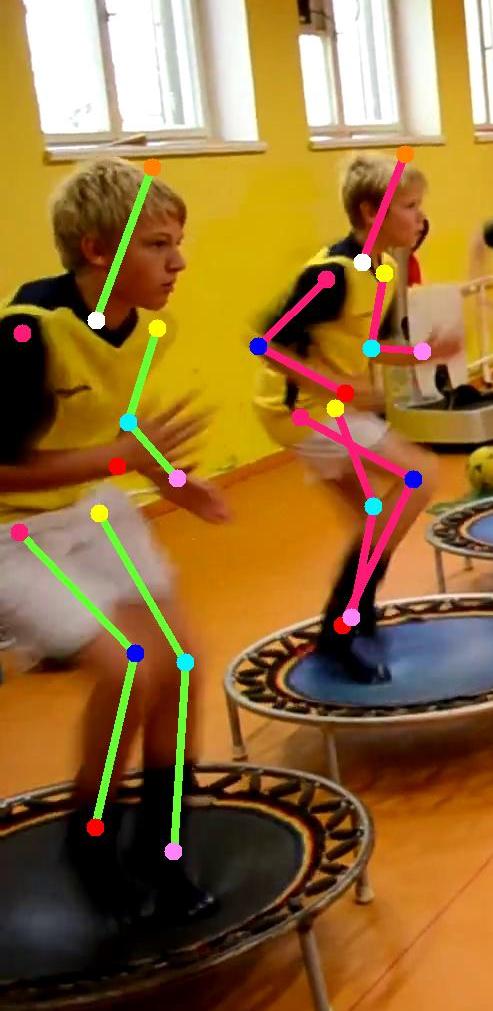} & {\footnotesize{}}\tabularnewline

GT Pose  & {\footnotesize{}} Initial Pose \cite{varadarajan2017greedy} & {\footnotesize{}} Refined Pose & GT Pose  & {\footnotesize{}} Initial Pose \cite{cao2017realtime} & {\footnotesize{}} Refined Pose \tabularnewline

\end{tabular}\hfill{}
\par\end{centering}
\vspace{0em}
\caption{\label{fig:qualitative-results-mpi} Qualitative results on the MPII Single-Pose dataset (\textbf{top}) and MPII Multi-Pose dataset (\textbf{bottom}). The blue circles denote the areas where the \texttt{PoseRefiner} brings significant improvement.
Our refinement method provides better localization for the challenging keypoint extremities (ankles and wrists), can remove confusions between symmetrical joint types (right ankle in \textbf{top right} and left ankle in \textbf{bottom left} figures) and can recover spurious joints (left wrist in \textbf{top left} figures) or missing joints (right hip in \textbf{bottom right} figures) by reasoning about the pose structure of the target person.
 }
\vspace{-1em}
\end{figure*}

\begin{enumerate}[wide, labelwidth=!, labelindent=0pt, label={(T\arabic*)}]

\item{Shift the coordinates of each joint by a displacement vector. 
The angle of the displacement vector is sampled uniformly from $[0, 2\pi]$. The length of the displacement vector is sampled with $90\%$ chance from $[0px, 25px]$
and $10\%$ chance from $[25px, 125px]$ to ensure both small and large displacements. 
In this way, the model is able to learn to do local refinements as well as to handle larger offsets of joints in spurious body configurations.} 

\item{Switch symmetric joints of the same person (e.g. replace left shoulder by right shoulder) - with probability $10\%$ per pair of joints. Such type of noise is a usual failure case of pose estimators, 
which occasionally confuse similarly looking left and right limbs or whether the joints are faced from the front or from the back.}

\item{Replace joints of a person by joints of the same/symmetric type of neighboring persons (e.g left hip of person A is replaced by the neighboring left/right hip of person B). 
Such synthesis is possible only when the pose estimation dataset contains multiple annotated people in the same image. If such a neighbor joint exists in a $75px$ vicinity, replacement is done with $30\%$ probability. 
This transformation models the pose estimation errors arising in crowded scenes, when limbs of different people are merged together.}

\item{Remove body joint with $30\%$ chance. This transformation helps to simulate the common missing joint error of body pose estimators, which is usually introduced by thresholding of low-confident keypoint detections.}

\end{enumerate}

%%%%%%%%%%%%%% Implementation Details %%%%%%%%%%%%%%%%%

\paragraph{Implementation Details}
We implement our system using the publicly available TensorFlow \cite{abadi2016tensorflow} framework. 

Following \cite{insafutdinov2016deepercut}, we rescale the input pose and image such that the reference height of a person is 340 px. 
The height of a person is estimated either from the scale of the ground truth head bounding box (if available, as in the MPII Single Person Dataset), 
or directly from the estimated input pose. We also crop 250 px in each direction around the bounding box of the input pose. 
This should standardize the input and minimize the searching space of the joints, while providing enough context to the \texttt{PoseRefiner}.

The input body pose estimate $P_{input}$ is encoded using one binary channel for each body joint. Each channel contains a circular blob of radius 15 px around the joint coordinate.
If there is no coordinate for the particular joint, the respective channel will be the null matrix. 
This encoding follows the encoding of the $P_{output}$ channels during training, which has been shown to work well for training strong body part detectors \cite{pishchulin2016deepcut,insafutdinov2016deepercut}.

Our training procedure contains a data augmentation step, which we employ for generating more training data. 
We apply random rescaling $\pm30\%$  and random flipping around the vertical axis.

When no pre-training is applied, we initialize the network with the weights of models trained on ImageNet \cite{deng2009imagenet}. 
For initialization of the extra convolutional filters corresponding to additional channels of $P_{input}$, we reuse the weights corresponding to RGB channels of $I$.

Optimization is done using stochastic gradient descent with $1$ image per batch, starting with learning rate $lr=0.005$ for one third of an epoch and continuing with $lr=0.02$ for $15$ epochs, $lr=0.002$ for $10$ epochs and $lr=0.001$ for $10$ 
other epochs. Training on the MPII dataset ($\approx 29k$ people) runs for 40 hours on one GPU \footnote {We use NVIDIA Tesla V100 GPU with 16 GB RAM}.

\section{\label{sec:Results} Results}
We now evaluate the proposed approach on the tasks of articulated single- and multi-person pose estimation, and articulated pose tracking.

\subsection{Experimental Setup}
We test our refinement network on three tasks involving human body pose estimation: single-person pose estimation, multi-person pose estimation and multi-person articulated tracking. 
In each of these tasks, we refine the predictions of several state-of-the-art methods by post-processing each initially estimated body pose using the \texttt{PoseRefiner}.

We experiment on four public challenges: MPII Human Pose \cite{andriluka20142d} ("Single-Person" and "Multi-Person") and PoseTrack \cite{andriluka2017posetrack} ("Single-Frame Multi-Person Pose Estimation" and "Multi-Person Articulated Tracking").

%%%%%%%%%%%%%%%%%% DATASETS %%%%%%%%%%%%%%%%%%
\paragraph{Datasets} For fair comparison with the methods whose prediction we refine, we follow the most common practices in choosing the datasets for training the \texttt{PoseRefiner}.

For the MPII Human Pose challenges, we train on MPII Human Pose only, whose training set contains $\approx 29k$ poses. 
For evaluation, we report results on the test set of MPII Human Pose, which includes $7,
247$ sufficiently separated poses used in the "Single-Person" challenge, as well as $4,485$ poses organized in groups of interacting people, 
used in the evaluation of the "Multi-Person" challenge. Although the protocol of the MPII Multi-Person task assumes that the location and the rough scale of each group of people is provided during test time, 
the \texttt{PoseRefiner} does not require any of this information.

In the case of the PoseTrack challenges, we pretrain on the COCO \cite{lin2014microsoft} train2017 set ($\approx 150k$ poses), then fine-tune on the MPII training set and afterwards on the PoseTrack training set. 
Pretraining is needed as the PoseTrack training set contains $\approx 61k$ poses, but only $2,437$ different identities, which do not cover a very high appearance variability. 
Since the set of joint types differs across datasets (MPII annotates $16$ keypoints, PoseTrack annotates $15$ and COCO annotates $17$), we use the PoseTrack set of body joints as reference 
and map all the other types to their closest PoseTrack joint type. The COCO dataset does not provide annotations for \textit{top-head} and \textit{bottom-head}, 
so we heuristically use the top most semantic segmentation vertex as the \textit{top-head} keypoint, and the midpoint between the \textit{nose} and the midpoint of the \textit{shoulders} as the \textit{bottom-head} keypoint. 
Similarly, MPII does not provide annotations for the \textit{nose} joint, so we use the midpoint between the \textit{bottom-head} and \textit{top-head} as a replacement. 
For evaluation, we report results on the PoseTrack validation set ($50$ videos, containing $18,996$ poses), which is publicly available.

%%%%%%%%%%%%%%%%%% EVALUATION METRICS %%%%%%%%%%%%%%%%%%
\paragraph{Evaluation metrics} For each task, we adopt the evaluation protocol proposed by the respective challenges.

On the MPII Human Pose (Single-Person) dataset, we report the Percentage of Correct Keypoints metric calculated with the matching threshold of half the length of the head segment ($PCK_h$@0.5), 
averaged across all joint types. We also report the Area Under the Curve measure (AUC), corresponding to the curve generated by $PCK_h$ measured over a range of percentages of the length of the head segment. 
Since none of the metrics is sensitive to false-positive joint detections, we do not remove non-confident predicted keypoints by thresholding them.

On the MPII Human Pose (Multi-Person) dataset, we report the mean Average Precision (mAP) based on the matching of body poses using $PCK_h$@0.5, following the evaluation kit of \cite{pishchulin2016deepcut}. 
This metric requires providing a confidence score for each detected body joint in addition to its location. Since the confidence scores provided by the \texttt{PoseRefiner} are in fact conditional probabilities of detection, 
we use the initial confidence scores (before refinement) that the input pose predictions come with. 

On the Single-Frame Multi-Person Pose Estimation task of PoseTrack, we report the same mAP metric as in MPII Human Pose, with the slight difference that the rough scale and location of people are not provided during test time, 
so the mAP evaluation in PoseTrack does not require it. 

On the Multi-Person Articulated Tracking task of PoseTrack, we calculate the Multiple Object Tracking Accuracy (MOTA) for each joint, and report mMOTA, averaged across all joints. 
This metric requires providing a track ID for each detected body pose, but no confidence score for joint detections. Since the metric is sensitive to false positive keypoints, we threshold the low confidence joints with the aim of 
removing incorrect predictions. 
We experimentally find that removing all joint detections with confidence scores less than $\tau = 0.7$ provides the best trade-off between the number of missed joints and the number of false positive joints, 
both penalized in the calculation of MOTA.

%%%%%%%%%%%%%%%%%% SINGLE PERSON %%%%%%%%%%%%%%%%%%
\subsection{\label{sec:Results-single-pose}Single-Person Pose Estimation}

The effect that the \texttt{PoseRefiner} has on the test set of MPII Single-Person is shown in Table~\ref{table:mpii-single-ok}. 
We refine various single person pose estimates given by different methods \cite{insafutdinov2016deepercut,gkioxari2016chained,carreira2016human,chen2017adversarial}, including the state of the art \cite{yang2017learning}.

\begin{figure*}
\begin{centering}
\setlength{\tabcolsep}{0em}
\renewcommand{\arraystretch}{0}
\par\end{centering}
\begin{centering}
\vspace{-1em}
\hfill{}%
\begin{tabular}{c@{\hskip 0.01in}c@{\hskip 0.01in}c@{\hskip 0.01in}c@{\hskip 0.01in}c@{\hskip 0.01in}c}

% First Sequence

\begin{sideways} \footnotesize{Input Poses} \cite{girdhar2017detect} \end{sideways} & {\footnotesize{}}
\includegraphics[width=0.19\textwidth]{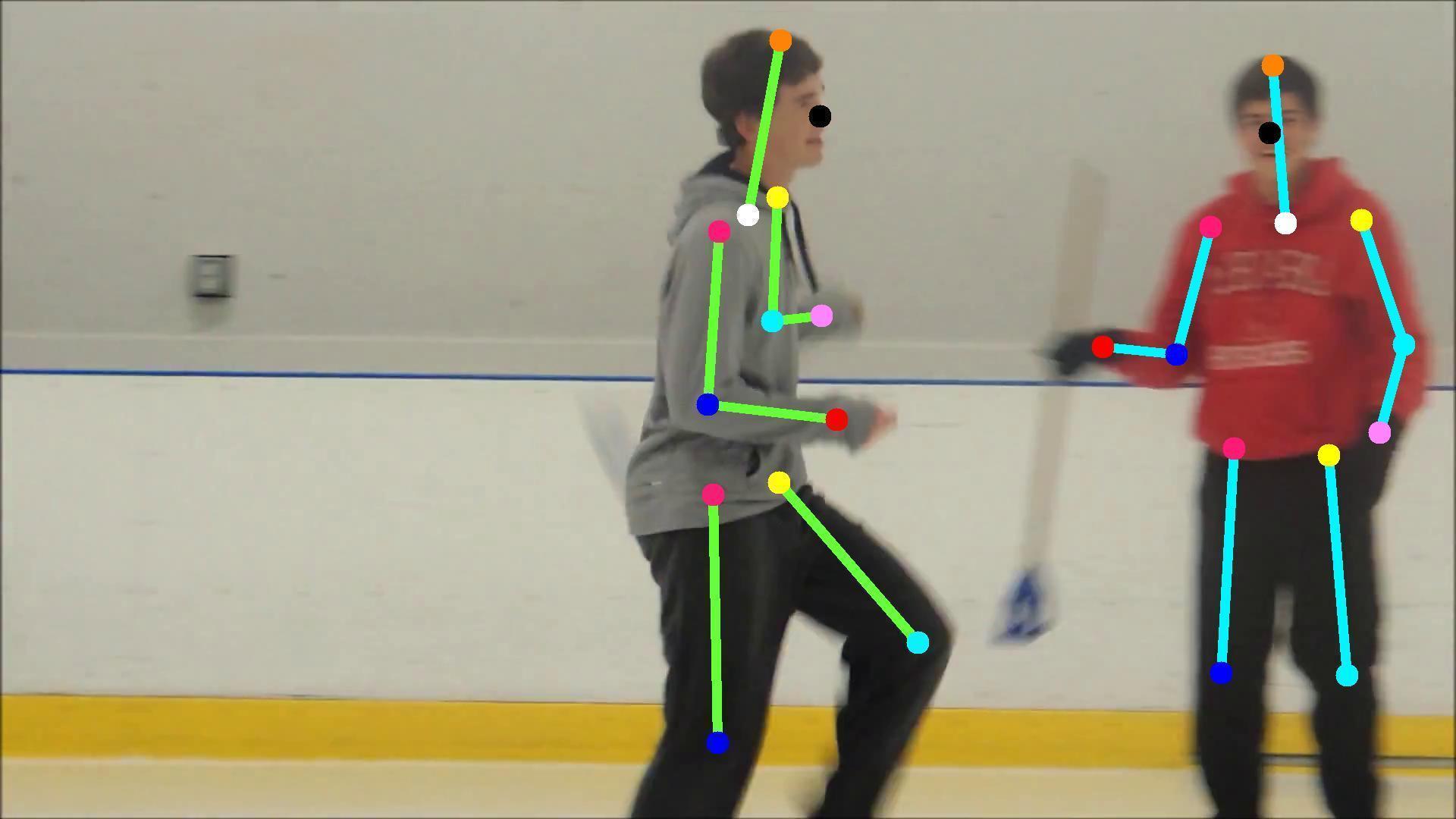} & {\footnotesize{}}
\includegraphics[width=0.19\textwidth]{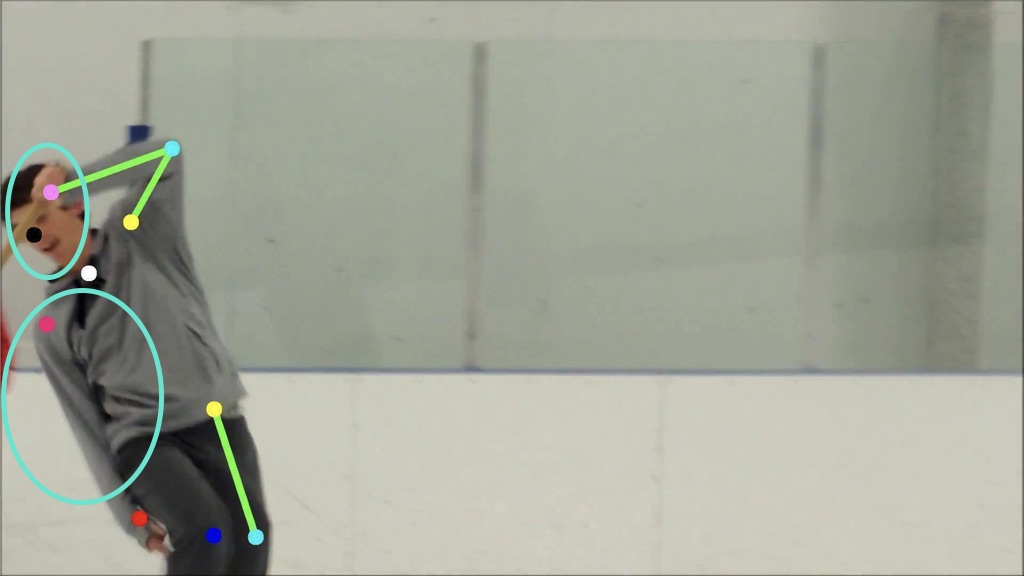} & {\footnotesize{}}
\includegraphics[width=0.19\textwidth]{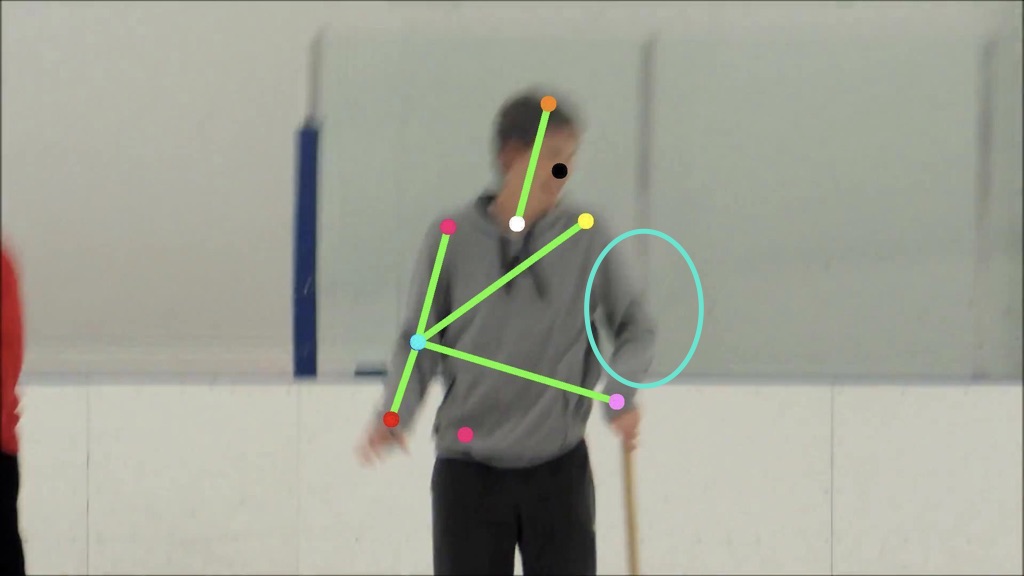} & {\footnotesize{}}
\includegraphics[width=0.19\textwidth]{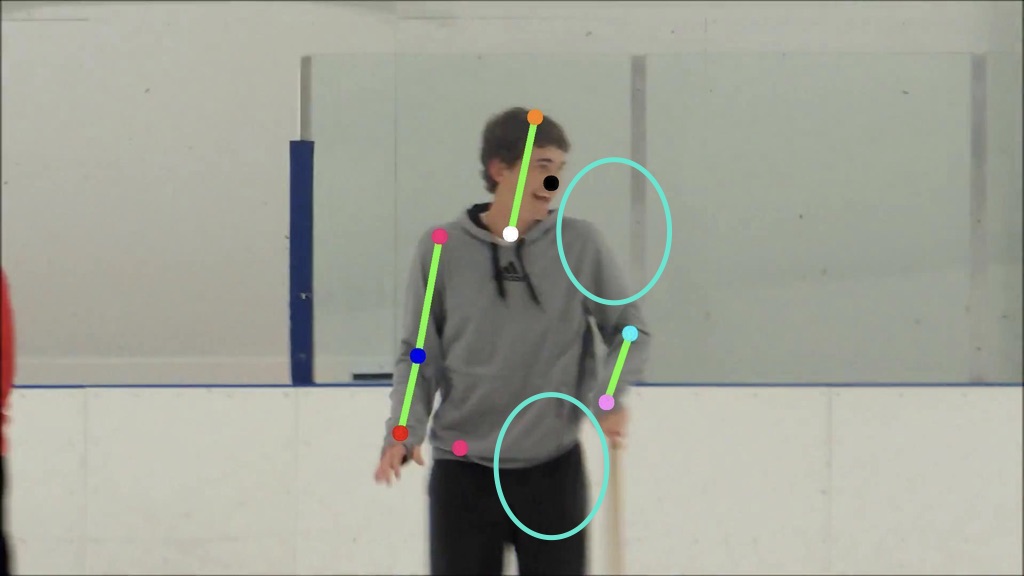} & {\footnotesize{}}
\includegraphics[width=0.19\textwidth]{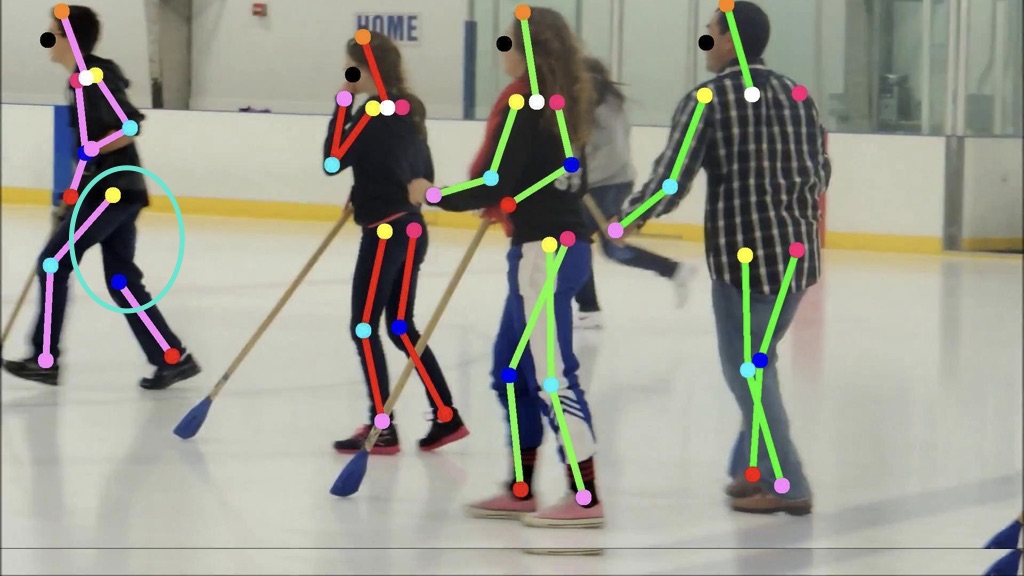} \tabularnewline

\begin{sideways} \footnotesize{Refined Poses} \end{sideways} & {\footnotesize{}}
\includegraphics[width=0.19\textwidth]{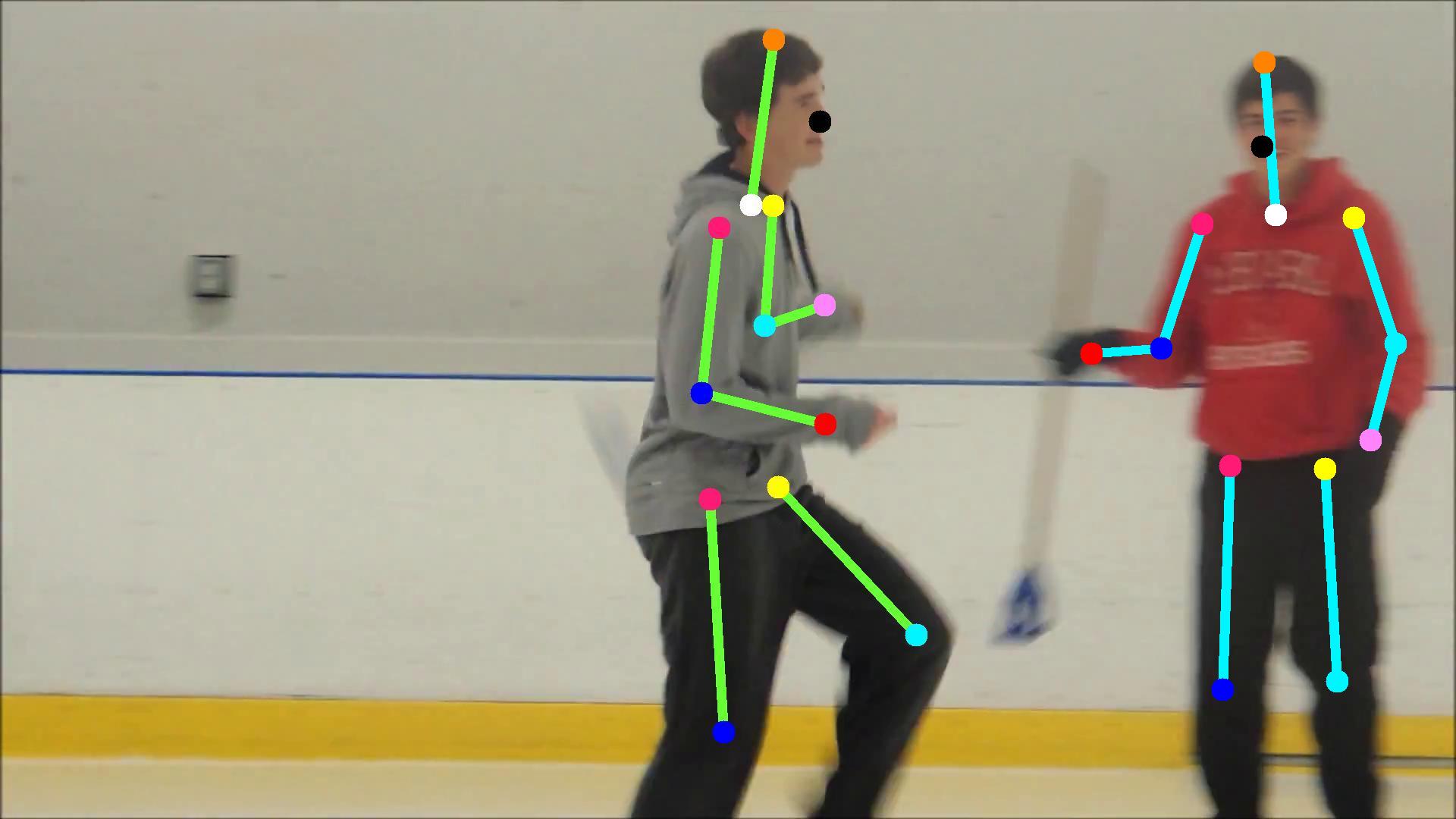} & {\footnotesize{}}
\includegraphics[width=0.19\textwidth]{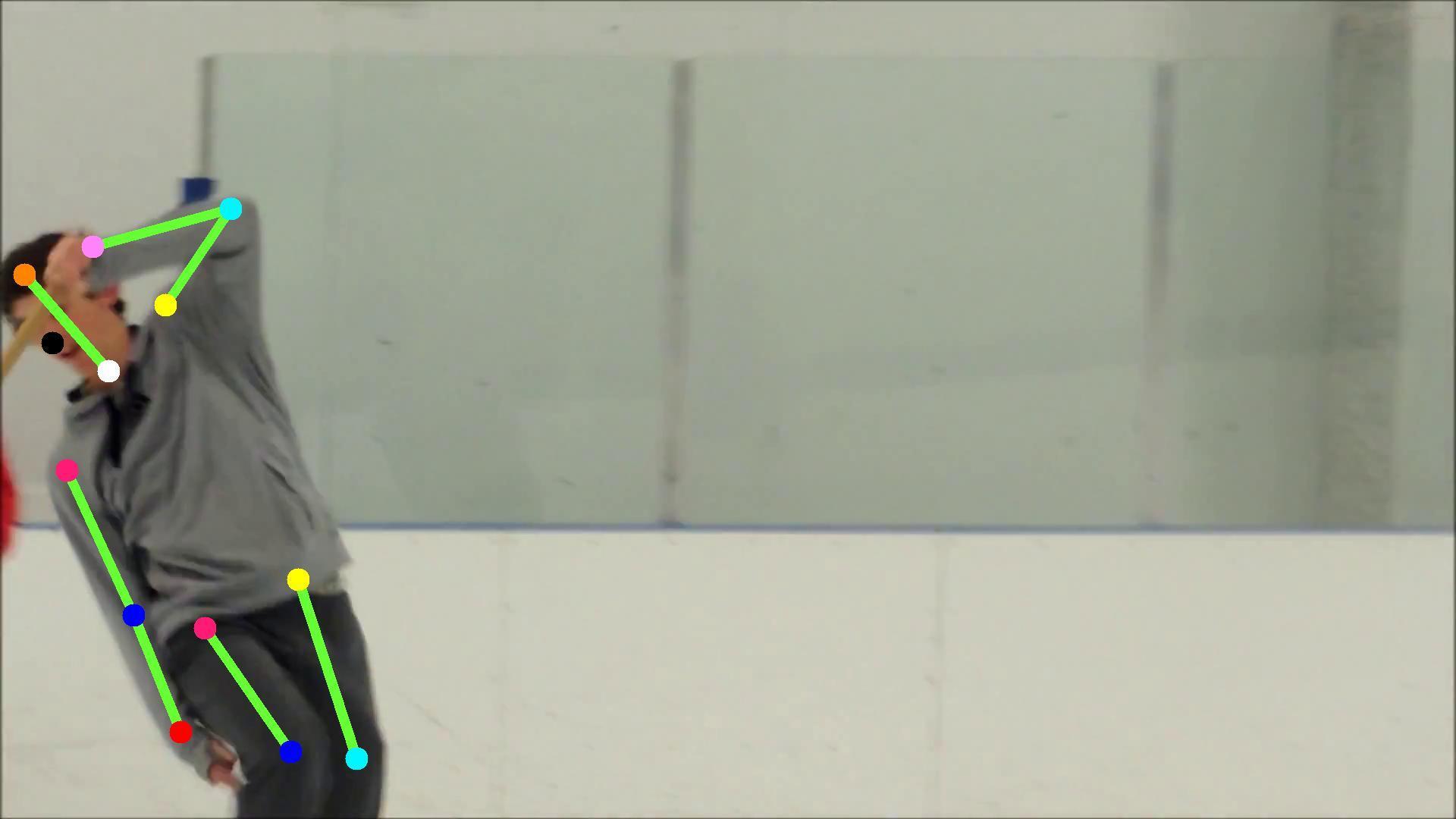} & {\footnotesize{}}
\includegraphics[width=0.19\textwidth]{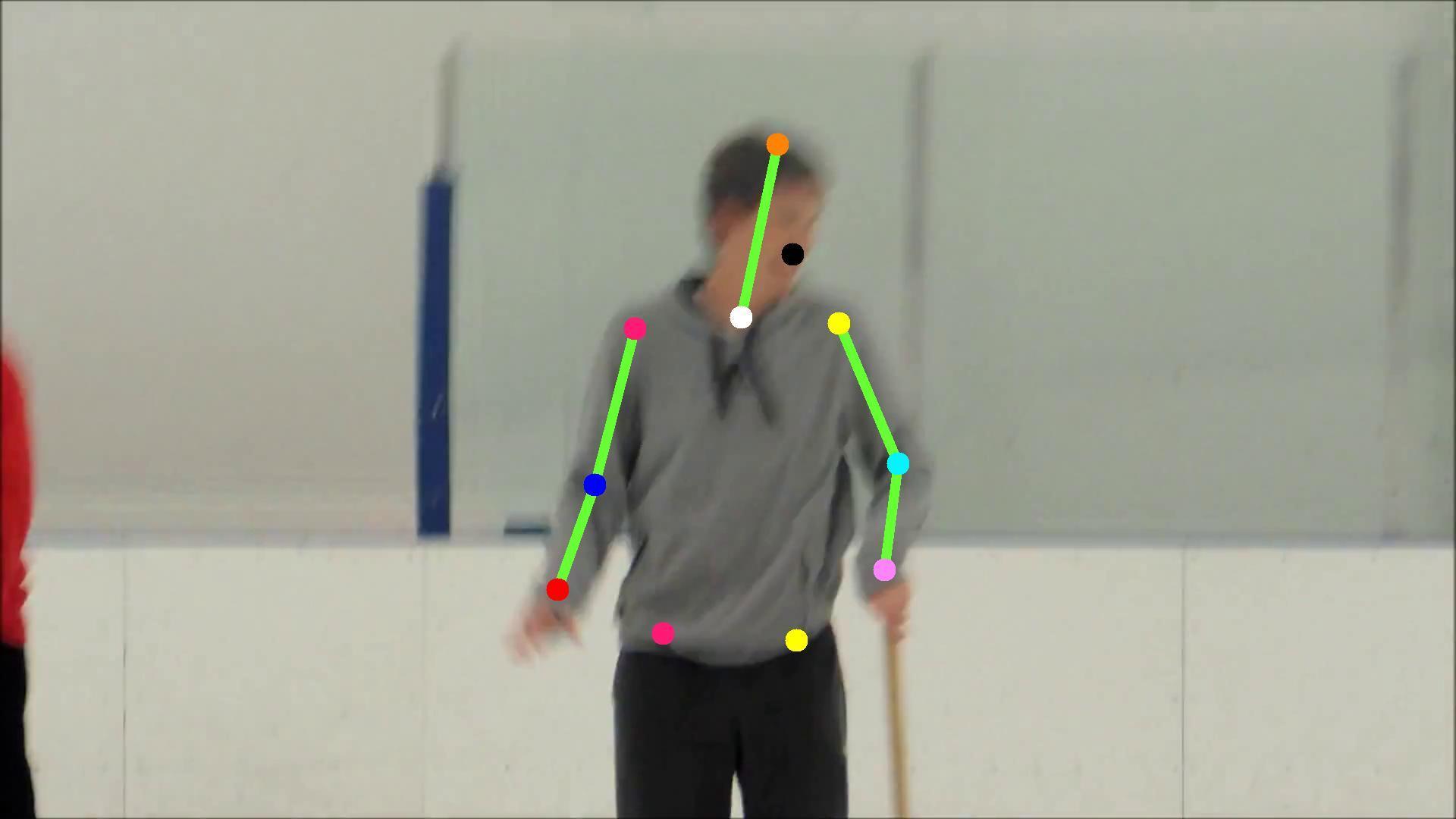} & {\footnotesize{}}
\includegraphics[width=0.19\textwidth]{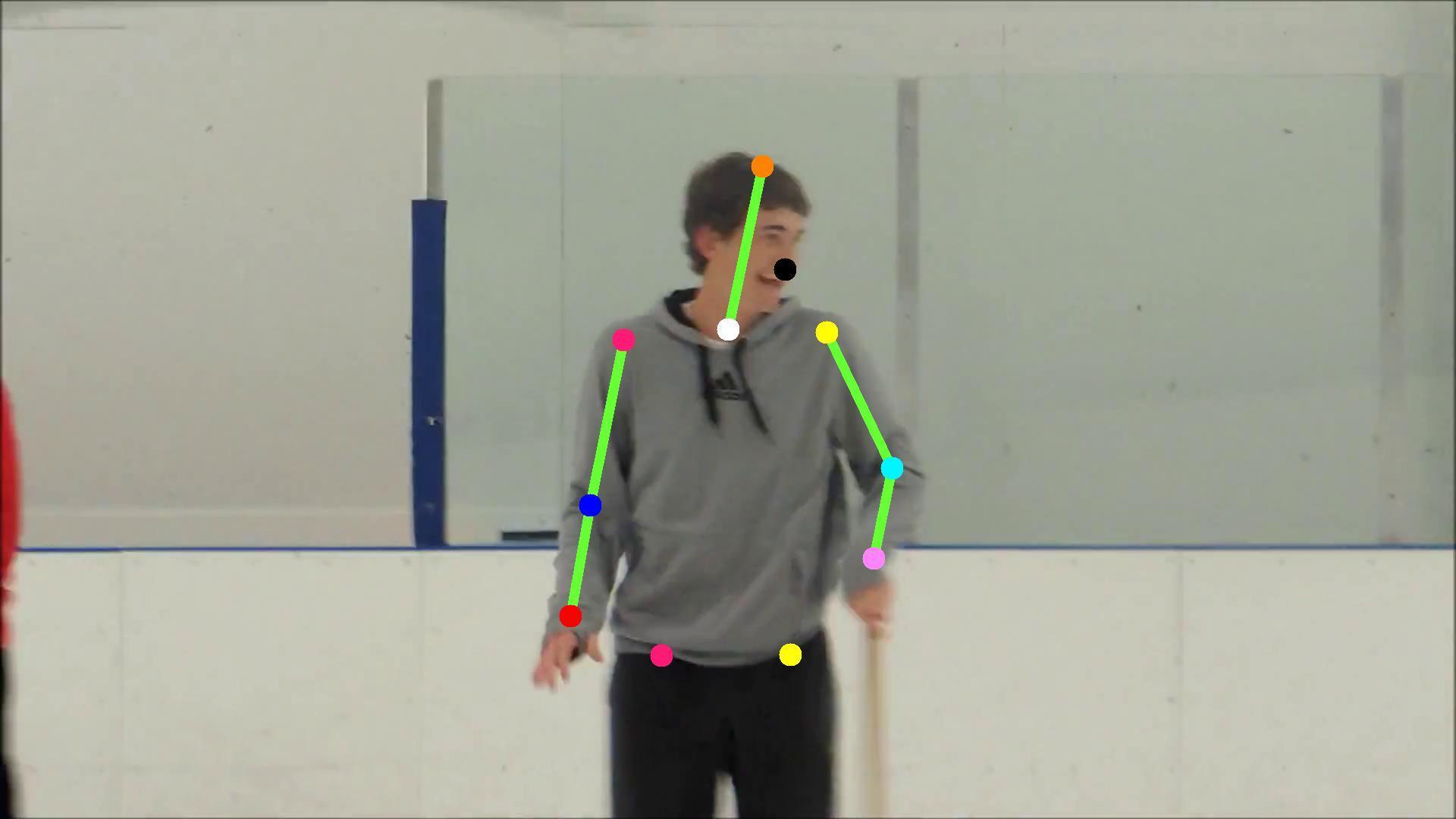} & {\footnotesize{}}
\includegraphics[width=0.19\textwidth]{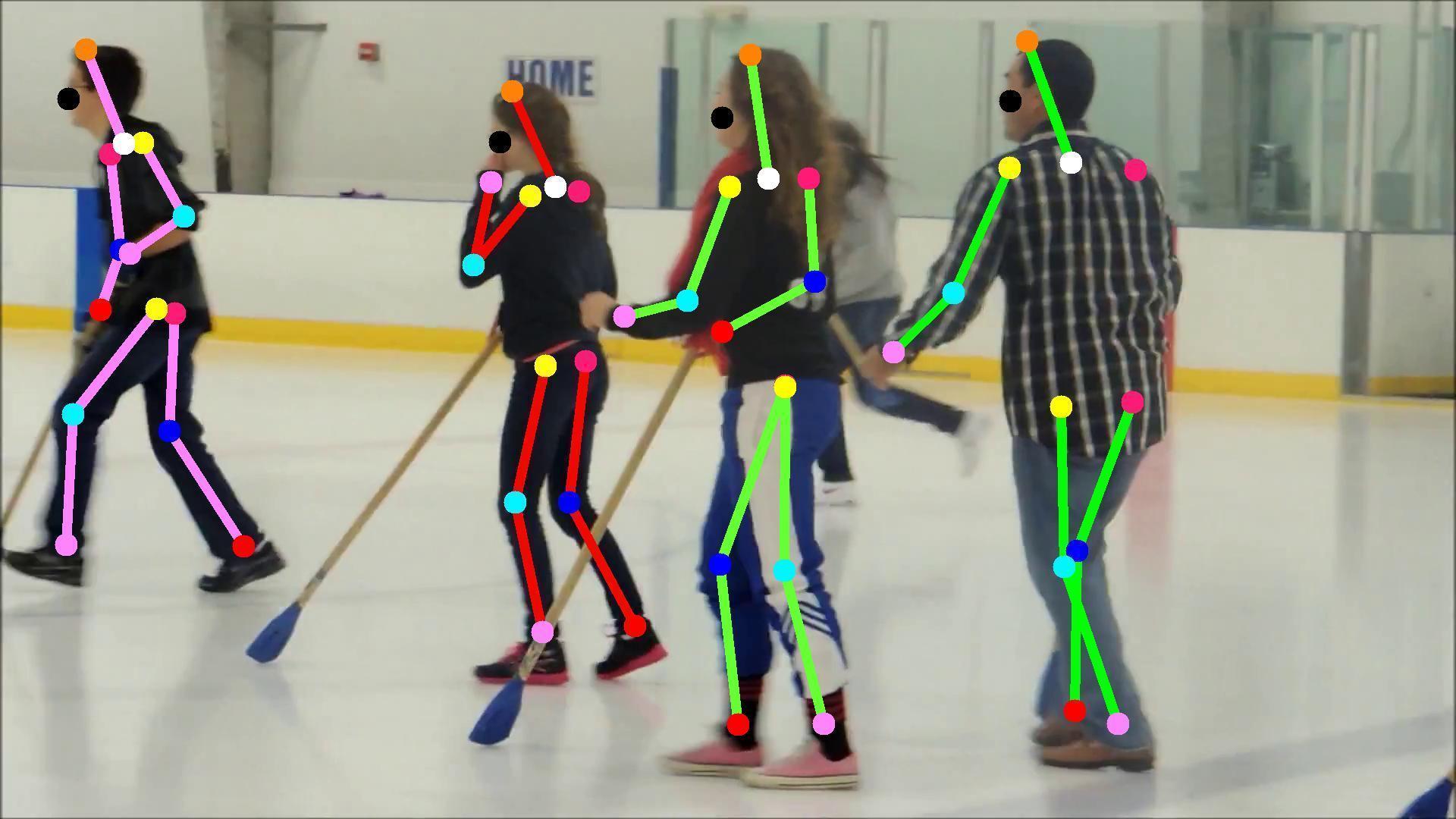} \tabularnewline

%%%%%%% Second Sequence %%%%%%%

\begin{sideways}\footnotesize{Input Poses} \cite{girdhar2017detect} \end{sideways} & {\footnotesize{}}
\includegraphics[width=0.19\textwidth]{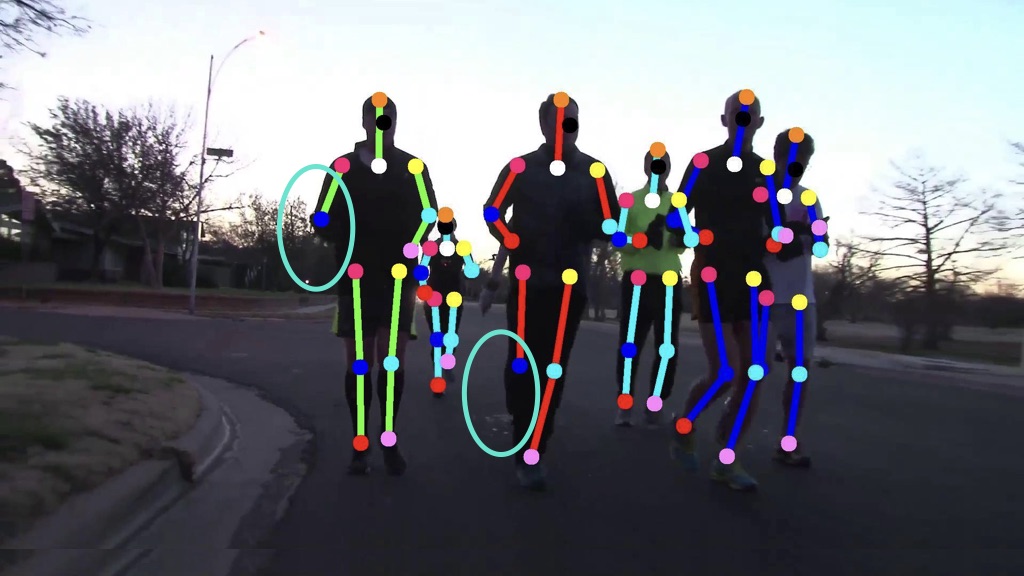} & {\footnotesize{}}
\includegraphics[width=0.19\textwidth]{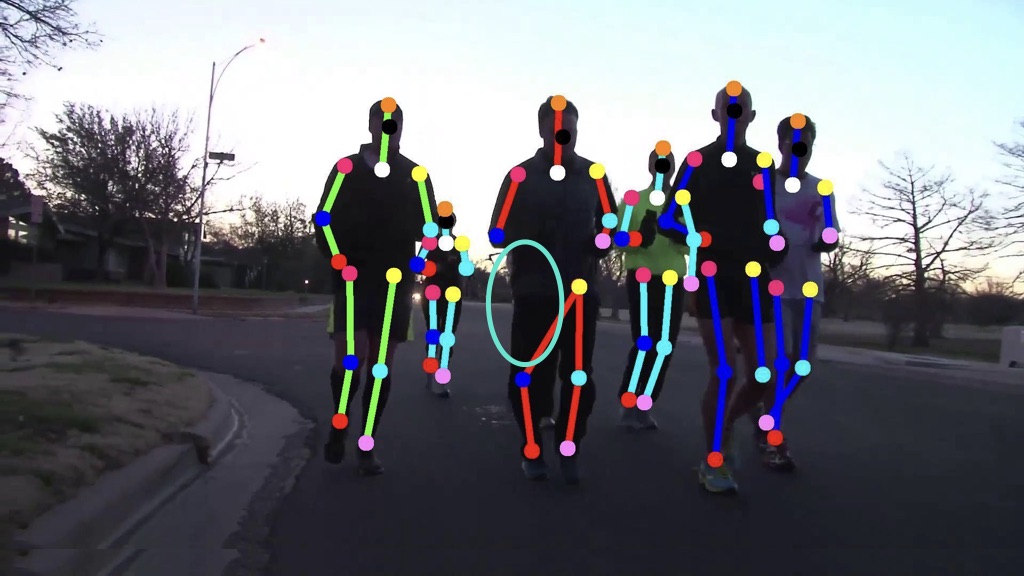} & {\footnotesize{}}
\includegraphics[width=0.19\textwidth]{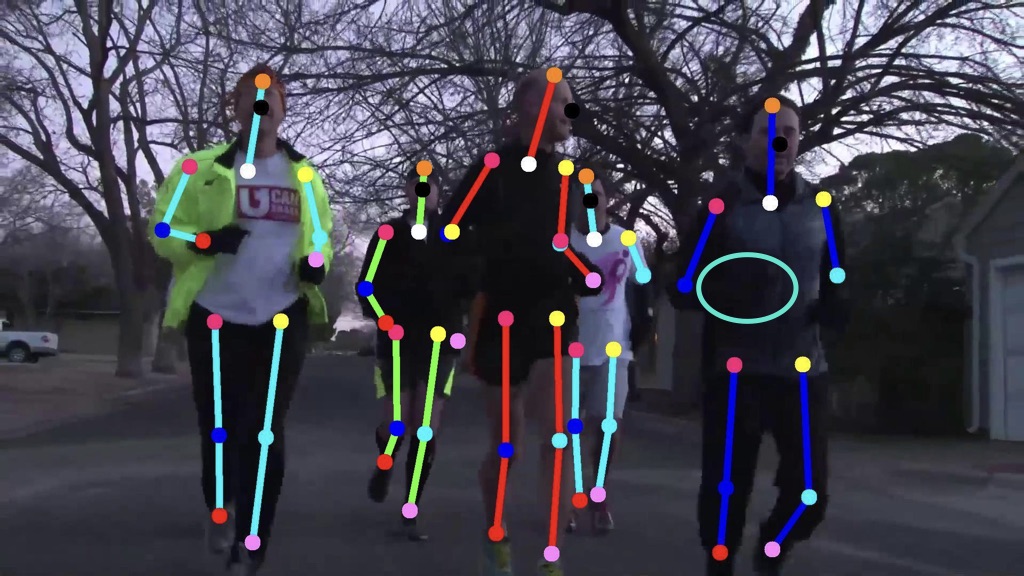} & {\footnotesize{}}
\includegraphics[width=0.19\textwidth]{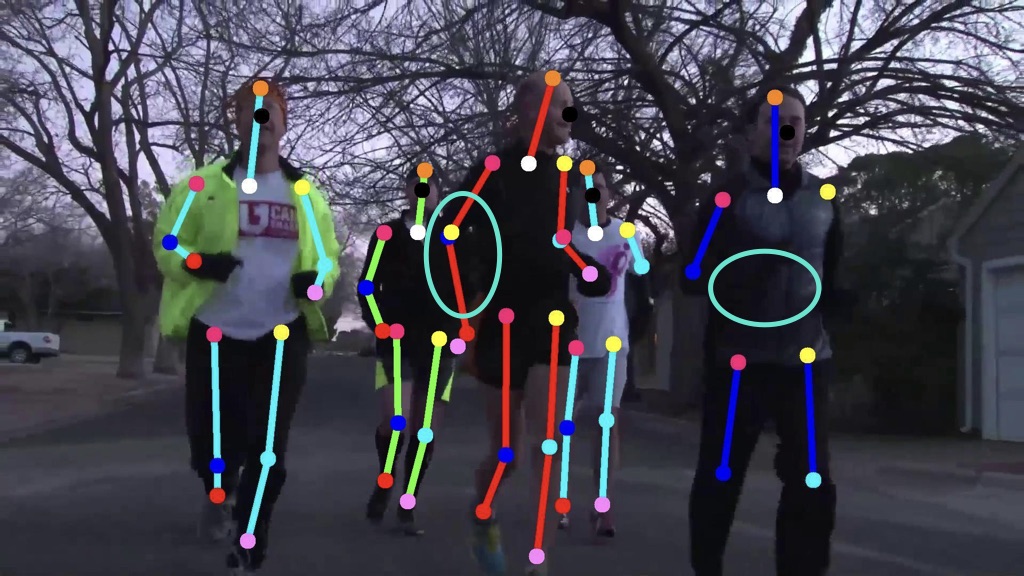} & {\footnotesize{}}
\includegraphics[width=0.19\textwidth]{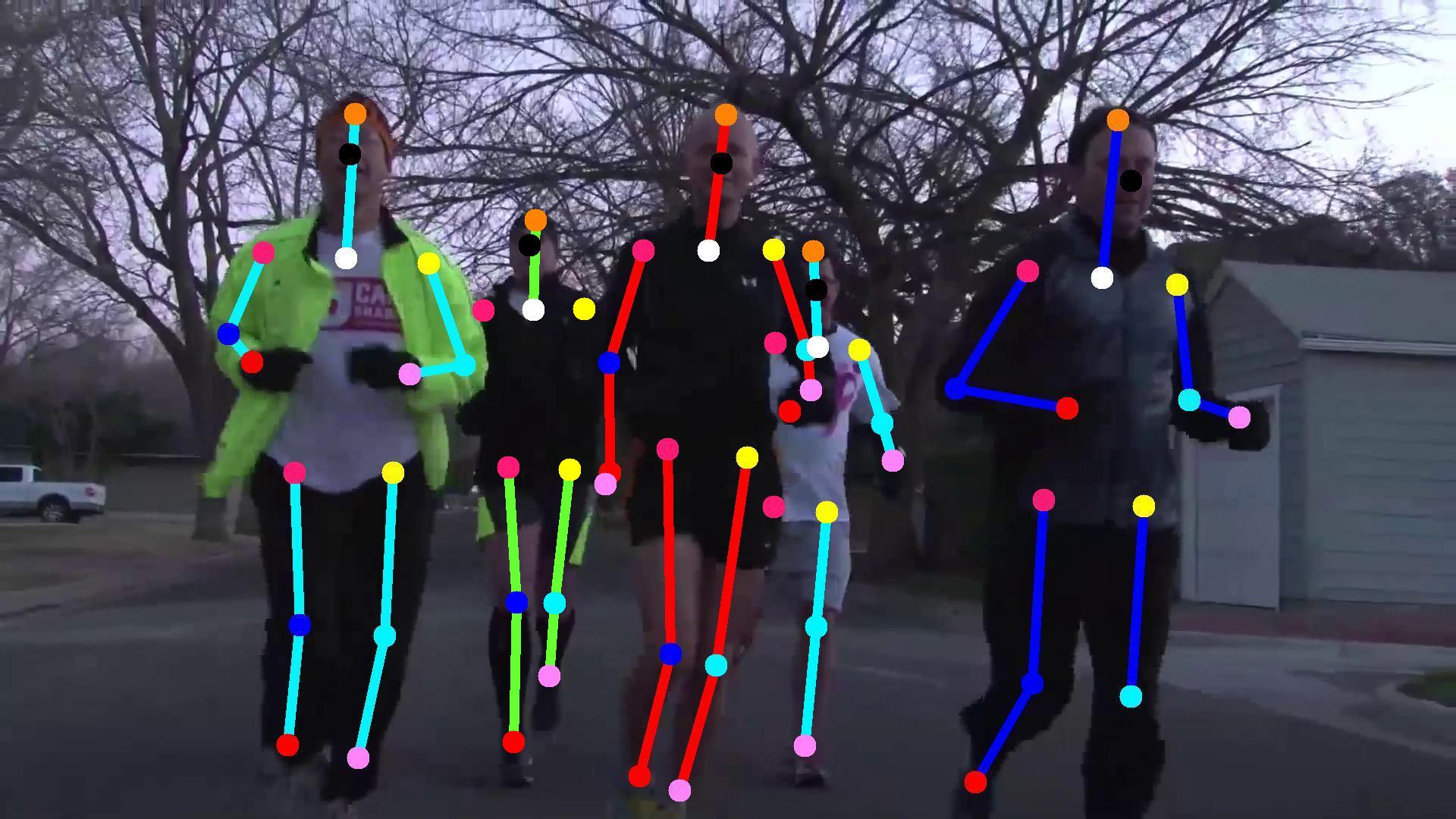} \tabularnewline

\begin{sideways} \footnotesize{Refined Poses} \end{sideways} & {\footnotesize{}}
\includegraphics[width=0.19\textwidth]{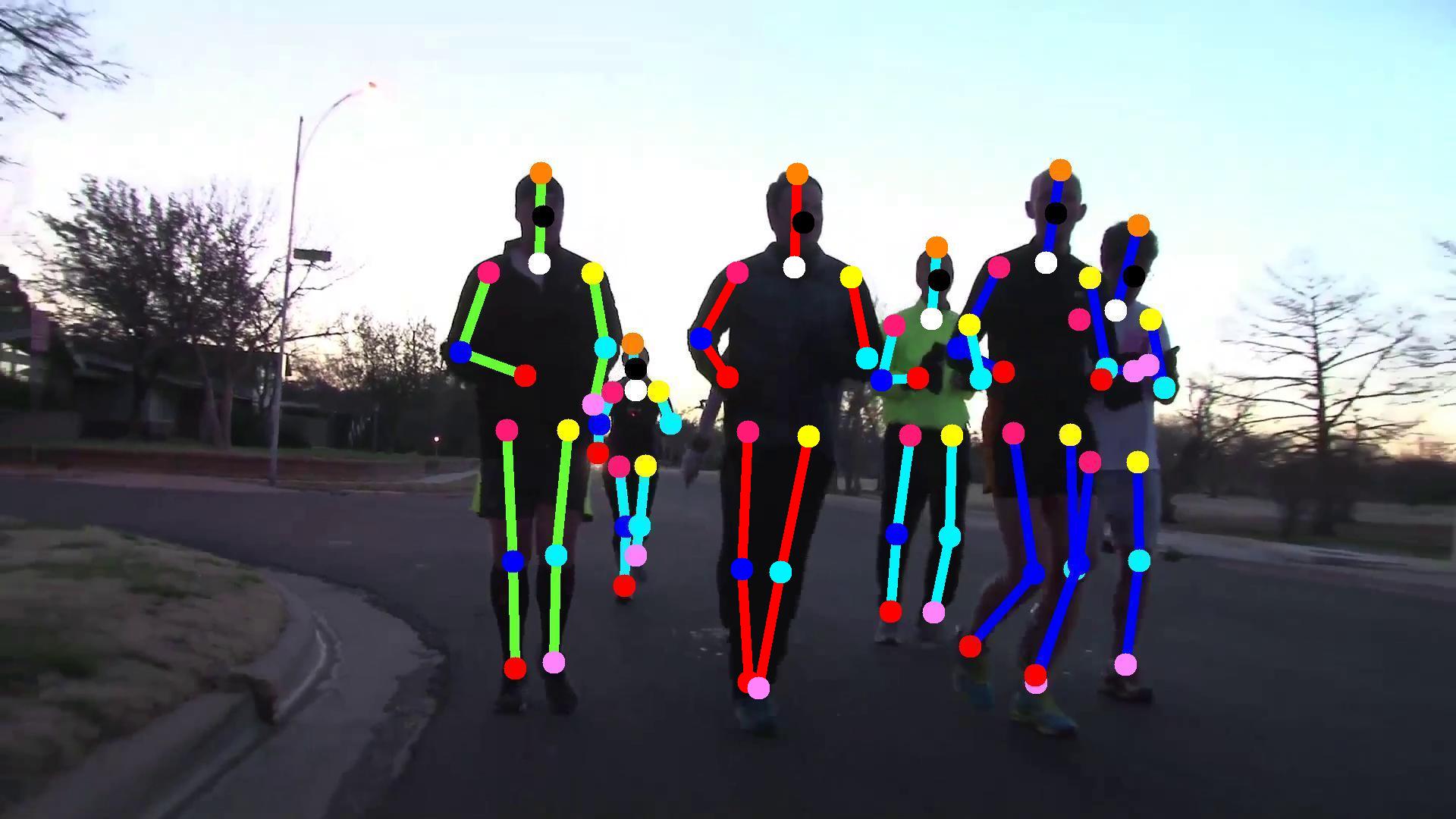} & {\footnotesize{}}
\includegraphics[width=0.19\textwidth]{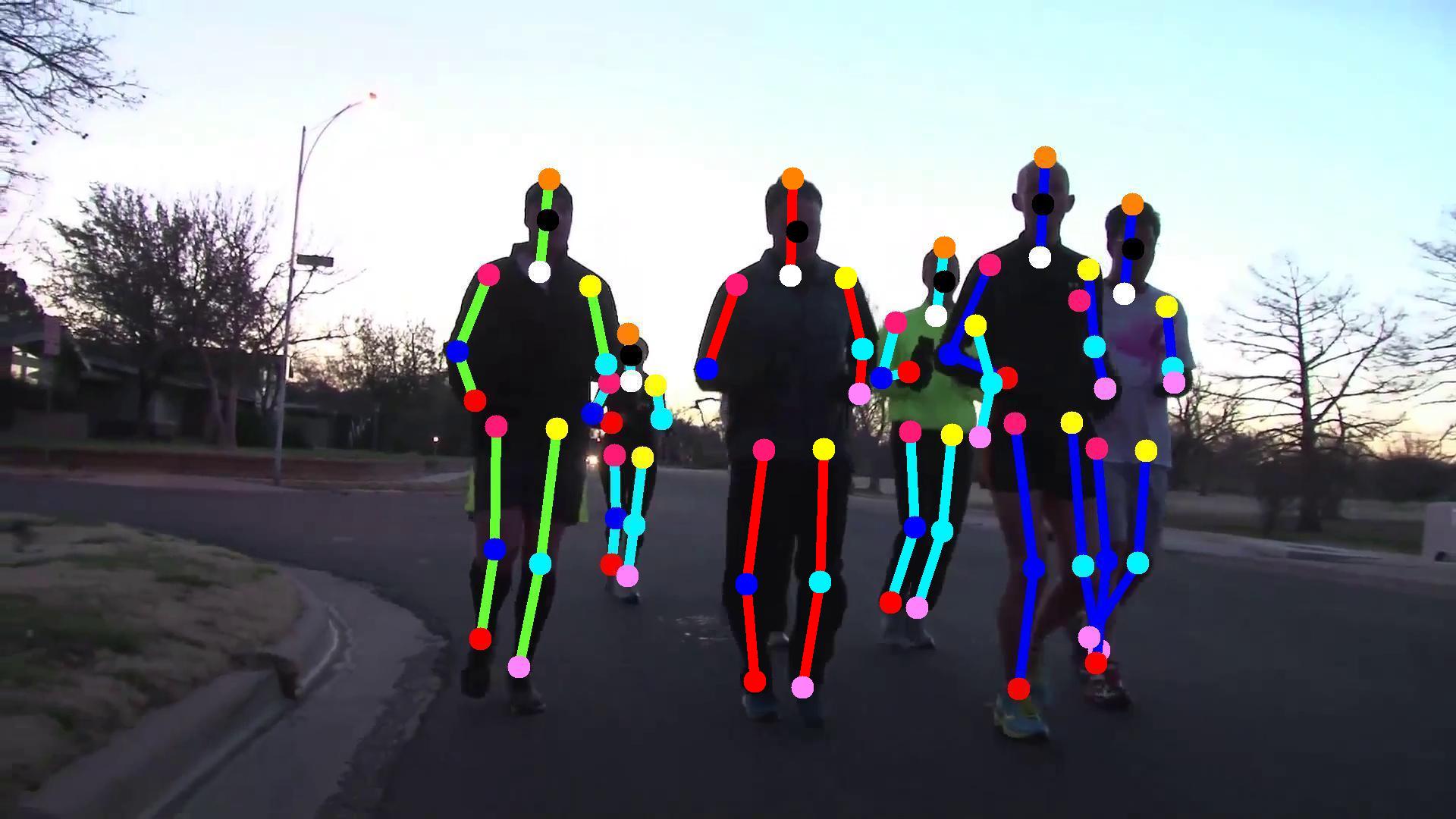} & {\footnotesize{}}
\includegraphics[width=0.19\textwidth]{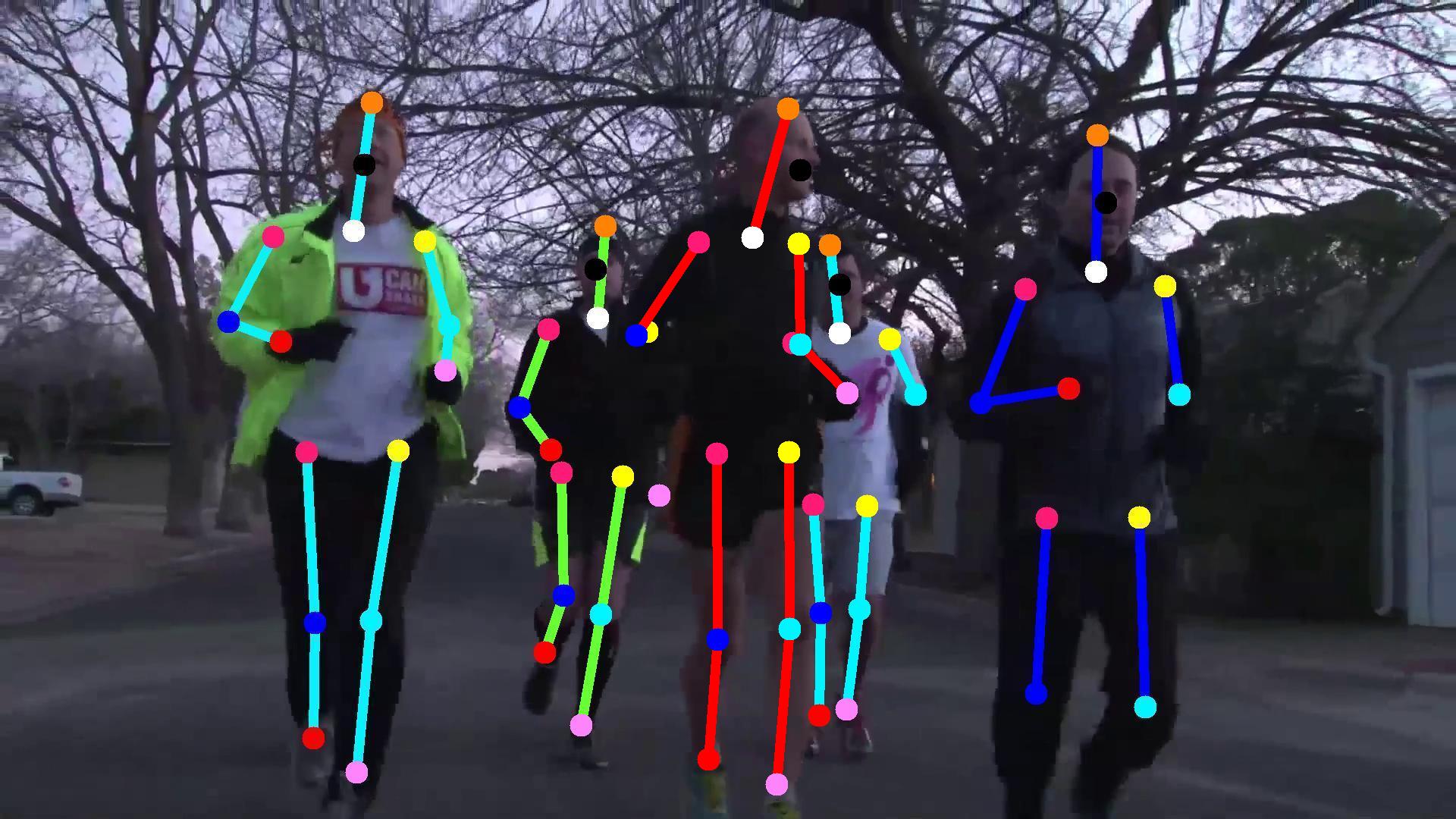} & {\footnotesize{}}
\includegraphics[width=0.19\textwidth]{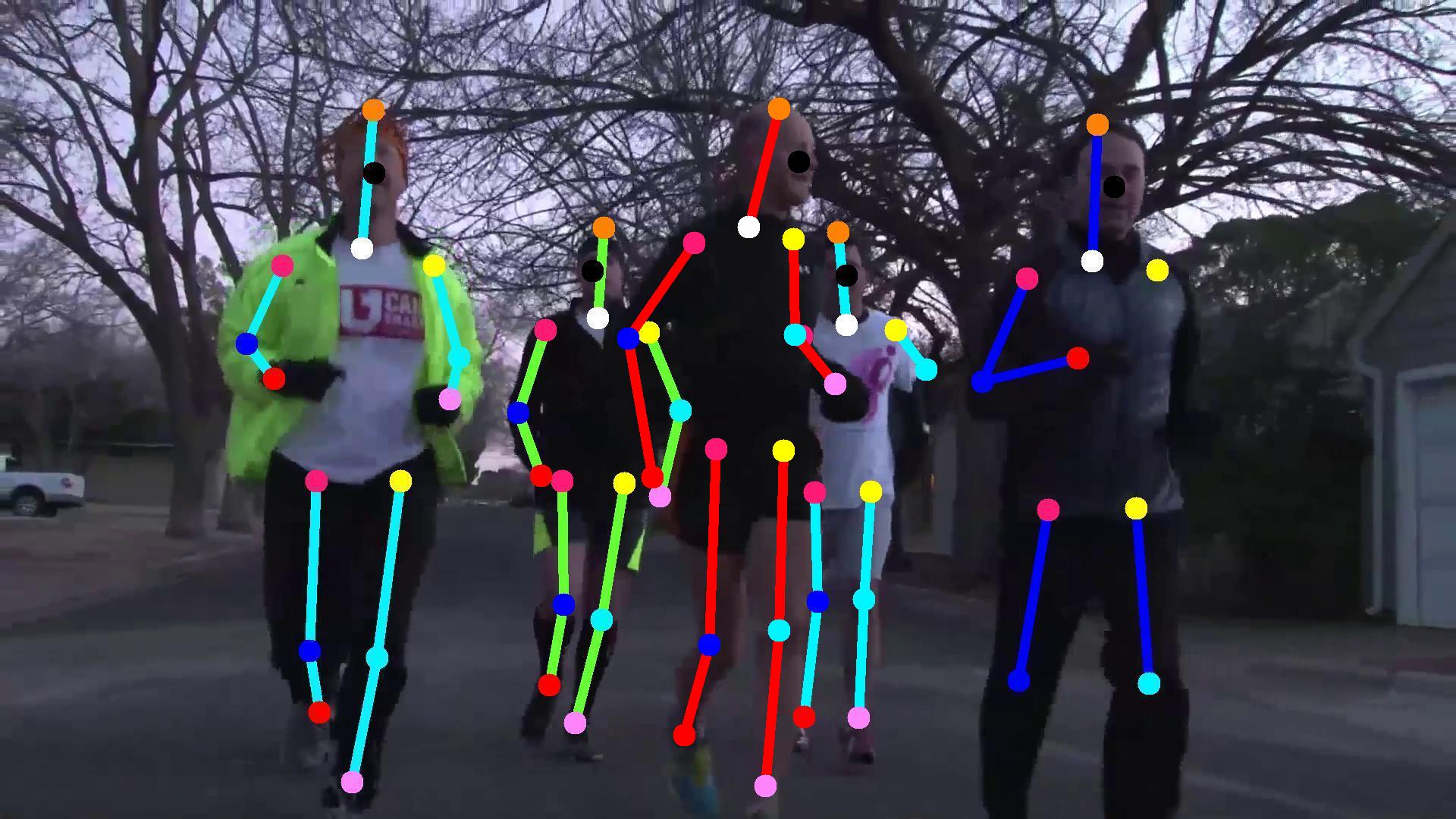} & {\footnotesize{}}
\includegraphics[width=0.19\textwidth]{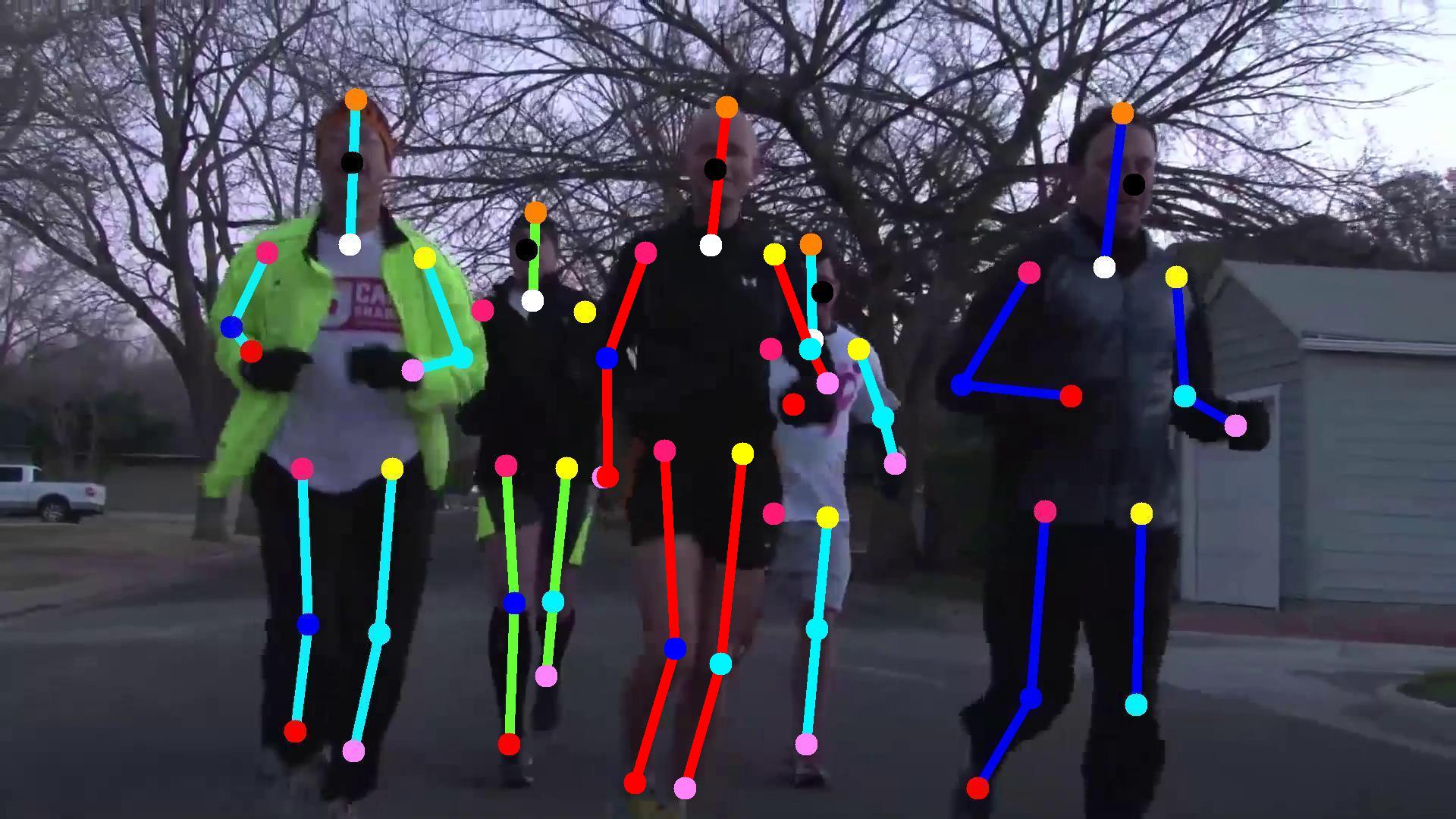} \tabularnewline

\end{tabular}\hfill{}
\par\end{centering}
\vspace{0em}
\caption{\label{fig:qualitative-results-posetrack} Qualitative results on the PoseTrack validation set, before and after applying the proposed refinement. The blue circles denote the areas where the proposed post-processing step brings significant improvement. The \texttt{PoseRefiner} recovers missing joints (e.g. right elbow and right hip in \textbf{top seq. - fr.2}, right wrist in \textbf{bottom seq. - fr.3}) and helps with confusions of symmetrical joints (left elbow in \textbf{top seq. - fr.3}, right hip in \textbf{bottom seq. - fr.2}). 
 }
\vspace{-1em}
\end{figure*}

One can notice that the performance of \cite{yang2017learning,chen2017adversarial} is already quite high, which motivates using less noise in the synthesis of the input pose for training the refinement model for these methods. 
Hence, we decrease the level of noise used in the generation of input poses during training by switching off all noise transformations, with the exception of (T1). 
We do not find it necessary to change the original set of noise transformations in any other experiment.

Using \texttt{PoseRefiner} as a post-processing step consistently increases the performance of each method, with the average improvement of $mPCK_h@0.5$ and AUC ranging from $0.3$ to $6.8$, while hurting neither metric. 
This shows the generality and effectiveness of our refinement method on single-person pose estimation, which already hints at its use in other more complex tasks involving keypoint detection, 
such as multi-person pose estimation in images and multi-person articulated tracking.

%%%%%%%%%%%%%%%%%%%%%%%%%%%%%%%%%%%%%%%%%%%%%%%%%%%%%%%%%%%%%%%%%%%%%%
\begin{table}
\setlength{\tabcolsep}{0.2em} 
\renewcommand{\arraystretch}{1.1}
\begin{center}
\begin{tabular}{lccc}
Method & $mPCK_h$@0.5 & AUC & $\Delta$ \\
\hline\hline

Pyramid Residual Module \cite{yang2017learning} & \textbf{92.0} & 64.2 & - \\
 ~~+ Refinement \footnotemark & \textbf{92.0}	& 64.7 & $+$0.3 \\
\hline

Adversarial PoseNet \cite{chen2017adversarial} & 91.9 & 61.6 & - \\
 ~~+ Refinement \footnotemark[\value{footnote}] & \textbf{92.1}	& 63.6 & $+$1.1 \\
\hline

DeeperCut \cite{insafutdinov2016deepercut} & 88.5 & 60.8 & -\\
 ~~+ Refinement & \textbf{89.1} & 62.3 & $+$1.0\\
\hline

Chained Predictions \cite{gkioxari2016chained} & 86.1 & 57.3 & -\\
 ~~+ Refinement & \textbf{88.0} & 61.2 & $+$2.9 \\
\hline

Iterative Error Feedback \cite{carreira2016human} & 81.3 & 49.1 & -\\
 ~~+ Refinement & \textbf{85.6} & 58.4 & $+$6.8\\
\hline

\end{tabular}
\end{center}
\caption{Effect of the proposed refinement over different pose estimation methods on the MPII Single-Person \cite{andriluka20142d} test set. $\Delta$ indicates the average improvement of mPCKh@0.5 and AUC after applying the pose refinement model.}
\label{table:mpii-single-ok}
\vspace{-1em}
\end{table}
\footnotetext{Using a refinement model trained with only (T1) transformations.}
%%%%%%%%%%%%%%%%%%%%%%%%%%%%%%%%%%%%%%%%%%%%%%%%%%%%%%%%%%%%%%%%%%%%%%
%%%%%%%%%%%%%%%%%%%%%%%%%%%%%%%%%%%%%%%%%%%%%%%%%%%%%%%%%%%%%%%%%%%%%%
\begin{table}
\setlength{\tabcolsep}{0.8em} 
\renewcommand{\arraystretch}{1.1}
\begin{center}
\begin{tabular}{lcc}
Method & mAP &  $\Delta$mAP\\
\hline\hline
 Associative Embedding \cite{newell2017associative} & 77.5 &  - \\
 ~~+ Refinement & \textbf{78.0} &  $+$0.5\\
\hline
Part Affinity Fields \cite{cao2017realtime} & 75.6 & - \\
 ~~+ Refinement & \textbf{76.9} &  $+$1.3\\
\hline
ArtTrack \cite{insafutdinov2017arttrack} & 74.2 &  -\\
 ~~+ Refinement & \textbf{75.1} &  $+$0.9 \\
\hline
Varadarajan et al., arXiv'17 \cite{varadarajan2017greedy} & 72.2 & - \\
 ~~+ Refinement & \textbf{75.1} &  $+$2.9\\
\hline

\end{tabular}
\end{center}
\caption{Effect of the proposed refinement over different methods on the MPII Multi-Person \cite{andriluka20142d} test set.  $\Delta$mAP indicates the improvement of mAP after applying the proposed pose refinement network.  }
\label{table:mpii-multi}
\vspace{-1em}
\end{table}
%%%%%%%%%%%%%%%%%%%%%%%%%%%%%%%%%%%%%%%%%%%%%%%%%%%%%%%%%%%%%%%%%%%%%%

We present the qualitative results on MPII Single Pose in Figure \ref{fig:qualitative-results-mpi}. Our refinement network is able to correct confusion between different joint types, recover from spurious or missing keypoints and provide better overall localization of joints.

%%%%%%%%%%%%%%%%%% MULTI PERSON %%%%%%%%%%%%%%%%%%
\subsection{\label{sec:Results-multi-pose}Multi-Person Pose Estimation}

Since the output of a multi-person pose estimator is a set of body poses in an image, we can use the \texttt{PoseRefiner} to perform error correction on each estimated pose, independently of the others.

Table~\ref{table:mpii-multi} shows the quantitative effect that the refinement post processing step has on several methods \cite{newell2017associative,cao2017realtime,insafutdinov2017arttrack,varadarajan2017greedy} 
applied on the MPII Multi-Person test set. It proves to help the overall performance of each system, including the best performing method \cite{newell2017associative} on this dataset, setting a new state of the art of $78.0$ mAP. 
Given that our system does not have any influence over non detected people/human body poses, the overall improvement (ranging from $0.5$ mAP to $2.9$ mAP) can be considered significant for the localization of joints.

Table~\ref{table:posetrack-multi} shows the results on the PoseTrack validation set. We refine the pose predictions of methods proposed for the Single-Frame Multi-Person Pose Estimation case \cite{ml_lab,insafutdinov2017arttrack,jin2017towards,girdhar2017detect}. 
They process images independently of each other and optimize the mAP metric. We again observe consistent improvements when employing the \texttt{PoseRefiner}, managing to increase the best reported performance on the dataset from $71.9$ mAP 
\cite{ml_lab} to $73.8$ mAP. 

In the case of ArtTrack \cite{insafutdinov2017arttrack}, which does not output a \textit{nose} joint, we remove the missing keypoint from the ground truth and from the evaluation procedure and report the obtained result ($68.6$ mAP) 
on the remaining subset of joints ($64.0
$ mAP evaluated on all joints). After post processing with the \texttt{PoseRefiner}, the \textit{nose} joint is recovered, and we report results for both evaluations: 
when removing the nose joint from the evaluation procedure ($70.0$ mAP) and when counting it into the evaluation ($69.7$ mAP). 
The fact that the difference between the two is small shows that the new \textit{nose} joint is recovered and nearly as well localized as the other joints. In addition, the overall performance after the refinement step is increasing ($68.6 \rightarrow 69.7$ mAP).
%%%%%%%%%%%%%%%%%%%%%%%%%%%%%%%%%%%%%%%%%%%%%%%%%%%%%%%%%%%%%%%%%%%%%%
\begin{table}
\setlength{\tabcolsep}{1em} 
\renewcommand{\arraystretch}{1.1}
\begin{center}
\begin{tabular}{lcc}
Method & mAP & $\Delta$mAP \\
\hline\hline
ML\_Lab \cite{ml_lab} & 71.9 & - \\
 ~~+ Refinement & \textbf{73.8} & $+$1.9 \\
\hline

ArtTrack \cite{insafutdinov2017arttrack} (best mAP) \footnotemark & 68.6 & - \\
~~+ Refinement (w/o nose) & \textbf{70.0} & $+$1.4 \\
~~+ Refinement (with nose) & 69.7 & $+$1.1 \\
\hline	

BUTD \cite{jin2017towards} (best mAP) & 67.8 & - \\
 ~~+ Refinement & \textbf{70.9} & $+$3.1\\
\hline

Detect-and-Track \cite{girdhar2017detect} & 60.4 & - \\
 ~~+ Refinement & \textbf{65.7} & $+$5.3\\
\hline
	
\end{tabular}
\end{center}
\caption{Effect of the proposed refinement on the PoseTrack \cite{andriluka2017posetrack} validation set, the Single-Frame Multi-Person Pose Estimation challenge. $\Delta$mAP indicates the improvement of mAP after applying the pose refinement model.}
\label{table:posetrack-multi}
\vspace{-1em}
\end{table}
\footnotetext{ArtTrack does not output a \textit{nose} joint, so the evaluation before refinement is performed without considering this joint. Our refinement network can recover the missing nose joint, 
leading to better performance ($68.6 \rightarrow 69.7$ mAP).}
%%%%%%%%%%%%%%%%%%%%%%%%%%%%%%%%%%%%%%%%%%%%%%%%%%%%%%%%%%%%%%%%%%%%%%

%%%%%%%%%%%%%%%%%% TRACKING %%%%%%%%%%%%%%%%%%
\subsection{\label{sec:Results-pose-tracking}Multi-Person Articulated Tracking}

Multi-Person Articulated Tracking involves detecting all people in each frame of a video, estimating their pose and linking their identities over time. 
We can therefore apply the \texttt{PoseRefiner} on each estimated pose independently of the others, while keeping the original identities of the detected people. 
Table~\ref{table:posetrack-track} shows the quantitative effect of the proposed refinement step on the PoseTrack validation set. Note that there are cases in which the results of the same method differ in 
Table~\ref{table:posetrack-multi} from Table~\ref{table:posetrack-track}, depending on which metric the method optimizes. Although the refinement only updates the coordinates of already detected body poses and no tracklet IDs are changed, 
the overall mMOTA improvement obtained by our system is significant (from $2.2$ to $4.9$ mMOTA). We show systematic improvement on every tracking result we refine, including the predictions of the method with the highest performance \cite{jin2017towards}. 
The state of the art is hence extended, reaching $58.4$ mMOTA on this benchmark. Similar to the Multi-Person Pose Estimation case, we recover the missing \textit{nose} joint on ArtTrack and manage to refine its overall 
tracking results by $3.9$ mMOTA. Qualitative results of multi-person articulated tracking are presented in Figure \ref{fig:qualitative-results-posetrack}.
%%%%%%%%%%%%%%%%%%%%%%%%%%%%%%%%%%%%%%%%%%%%%%%%%%%%%%%%%%%%%%%%%%%%%%
\begin{table}
\setlength{\tabcolsep}{0.25em} 
\renewcommand{\arraystretch}{1.1}
\begin{center}
\begin{tabular}{lccc}
Method & mAP & mMOTA & $\Delta$mMOTA\\
\hline\hline

BUTD \cite{jin2017towards} (best mMOTA) & 62.5 & 56.0 & -\\
 ~~+ Refinement & 64.3 & \textbf{58.4} & $+$2.4\\
\hline

Detect-and-Track \cite{girdhar2017detect} & 60.4 & 55.1 & - \\
 ~~+ Refinement & 64.1 & \textbf{57.3} & $+$2.2\\
\hline

ArtTrack \cite{insafutdinov2017arttrack} (best mMOTA) \footnotemark & 66.7 & 50.2 & -\\
 ~~+ Refinement (w/o nose) & 66.5 & 53.3 & $+$3.1\\
 ~~+ Refinement (with nose) & 67.0 & \textbf{54.1} & $+$3.9 \\
\hline

ML\_Lab \cite{ml_lab} & 71.9 & 48.6 & -\\
 ~~+ Refinement & 70.1 & \textbf{53.5} & $+$4.9 \\ 
\hline

\end{tabular}
\end{center}
\caption{Effect of the proposed refinement on the PoseTrack \cite{andriluka2017posetrack} validation set, the Multi-Person Articulated Tracking challenge.
$\Delta$mMOTA indicates the improvement of mMOTA after applying the pose refinement model. 
}
\label{table:posetrack-track}
\vspace{-1em}
\end{table}
\footnotetext{Although ArtTrack does not output a \textit{nose} joint, our refinement network can recover the missing nose joint, while improving overall performance ($50.2 \rightarrow 54.1$ mMOTA).}
%%%%%%%%%%%%%%%%%%%%%%%%%%%%%%%%%%%%%%%%%%%%%%%%%%%%%%%%%%%%%%%%%%%%%%

\section{\label{sec:Conclusion} Conclusion}
In this work we proposed a human pose refinement network which can be applied over a body pose estimate derived from any human pose estimation approach.
In comparison to other refinement techniques, our approach provides a simpler solution by directly generating the refined body pose from the initial pose prediction in one forward pass, 
exploiting the dependencies between the input and output spaces.
We report consistent improvement of our model applied over state-of-the-art methods across different datasets and tasks, highlighting its effectiveness and generality.
Our experiments show that even top performing methods can benefit from the proposed refinement step.  
With our refinement network we improve the best reported results on MPII Human Pose and PoseTrack datasets for multi-person pose estimation and pose tracking tasks.

{\small
\bibliographystyle{ieee}
\bibliography{egbib}
}

\end{document}